\documentclass[a4paper]{article}       
\usepackage{graphicx}
\usepackage{amsmath,amsfonts,amssymb}
\usepackage{subcaption}
\usepackage{multicol}
\usepackage{microtype}
\usepackage{mathtools}
\usepackage{url}
\usepackage{times}
\usepackage{setspace}
\usepackage{tabularx}
\usepackage{xspace}
\usepackage{color,soul}
\usepackage{fancyvrb}
\usepackage{listings}
\usepackage{algorithm2e}
\usepackage{authblk}
\newcommand{\XTT}{\textsc{XTT2}\xspace}
\newcommand{\KNAC}{\textsc{KnAC}\xspace}
\newcommand{\HRTD}{\mbox{\textsc{HeaRTDroid}}\xspace}
\graphicspath{{./pix/}}
\usepackage{mathptmx}      
\begin{document}

\title{%
\KNAC: an approach for enhancing cluster analysis with background knowledge and explanations.
}


\author[1]{Szymon Bobek}
\author[2]{Michał Kuk}
\author[1]{Jakub Brzegowski}
\author[2]{Edyta Brzychczy}
\author[1]{Grzegorz J. Nalepa}

\affil[1]{Jagiellonian Human-Centered Artificial Intelligence Laboratory (JAHCAI) and \\
              Institute of Applied Computer Science \\
              Jagiellonian University, 31-007 Krak\'ow, Poland}
              
\affil[2]{AGH University of Science and Technology, Krak\'ow, Poland}

\date{Accepted: 2022.10.30 to Applied Intelligence}

\maketitle

\begin{abstract}
Pattern discovery in multidimensional data sets has been the subject of research for decades.
There exists a wide spectrum of clustering algorithms that can be used for this purpose.
However, their practical applications share a common post-clustering phase, which concerns expert-based interpretation and analysis of the obtained results.
We argue that this can be the bottleneck in the process, especially in cases where domain knowledge exists prior to clustering.
Such a situation requires not only a proper analysis of automatically discovered clusters but also conformance checking with existing knowledge.
In this work, we present Knowledge Augmented Clustering (\KNAC). Its main goal is to confront expert-based labelling with automated clustering for the sake of updating and refining the former.
Our solution is not restricted to any existing clustering algorithm.
Instead, \KNAC can serve as an augmentation of an arbitrary clustering algorithm, making the approach robust and a model-agnostic improvement of any state-of-the-art clustering method.
We demonstrate the feasibility of our method on artificially, reproducible examples and in a real life use case scenario.
In both cases, we achieved better results than classic clustering algorithms without augmentation.

\end{abstract}

\section{Introduction}
\label{sec:intro}

Cluster discovery from highly dimensional data is an important area of research in the fields of data mining (DM) and  machine learning (ML).
Most of the research in this area is focused on unsupervised approaches that generate clusters which have to be carefully analysed by experts to obtain their semantic interpretation.
In almost all cases, the responsibility for the final evaluation and assessment of the results is left to the domain experts.
This fact is exploited by some of the approaches that utilise prior knowledge about possible clusters in supervised  or semi-supervised algorithms to improve the performance or quality of the results.
However, we argue that this exposes the bottleneck in the practical application of the above-mentioned solutions as the expert background knowledge can be altered with the results of the clustering.
This invalidates prior knowledge used for clustering and complicates the analysis of the results, and
may happen when unexpected patterns are discovered in the data that are  in contradiction with the existing knowledge or that extend this knowledge.

This phenomenon is especially visible in applications of DM in Industry 4.0~\cite{liu2019human}, where data is gathered from a process that is usually well known and described in its generic form, however, data exposes more patterns than was originally perceived by experts. 
The difficulty lies in the automation of the data mining process, and the conformance checking between the theoretical knowledge and the data delivered by the system.
On the level of industrial or business processes, such conformance checking can be done with tools designed to work on such a high level of abstraction~\cite{confchecking2008}.
However, on the low level of multidimensional sensory data, where no process is yet visible, atomic pattern discovery is the main objective, which can be achieved with a variety of available clustering algorithms.
These algorithms do not inherently provide any tools to support the task of confronting discovered clusters with the existing knowledge.

\begin{figure}
\centering
\includegraphics[width=1\textwidth]{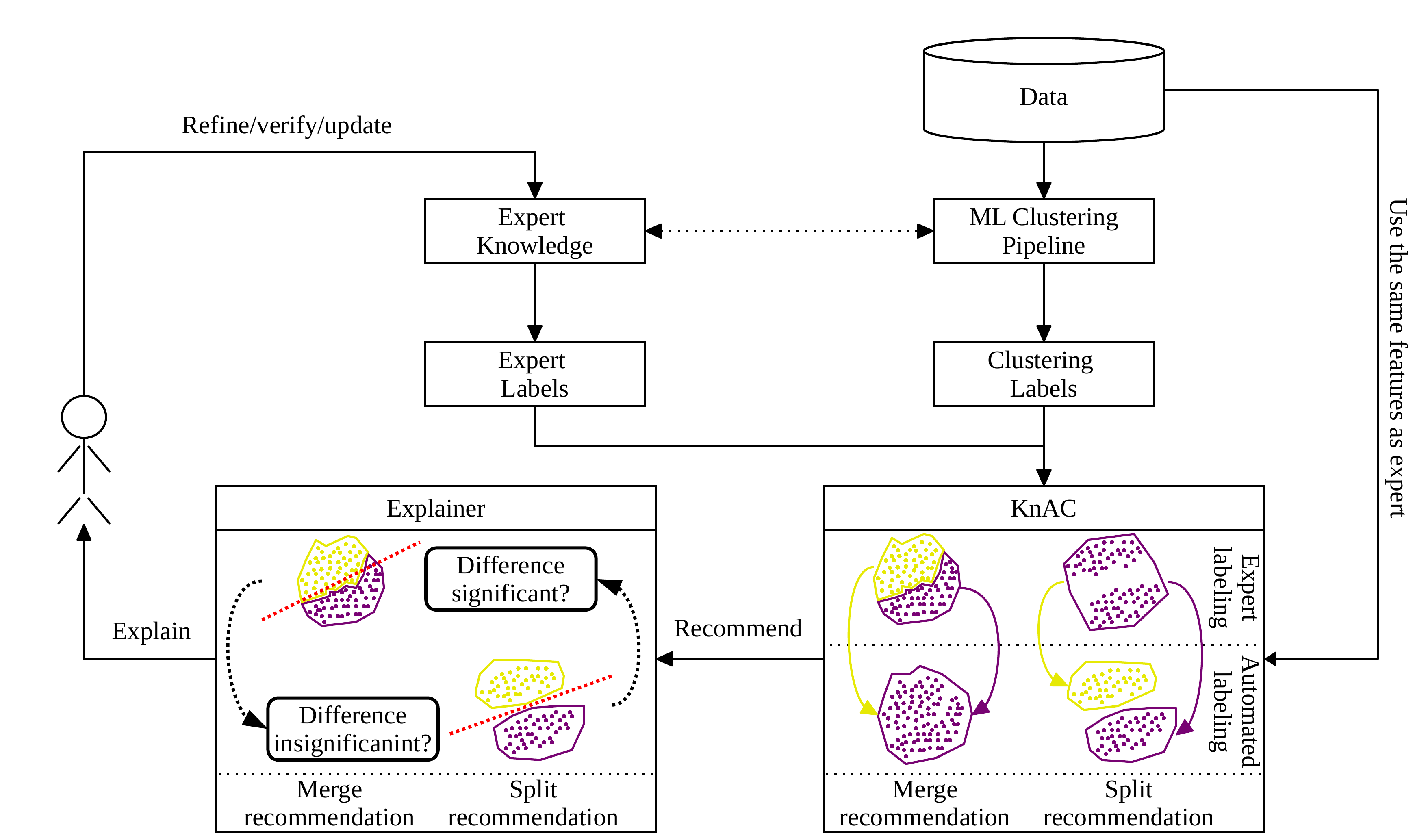}
\caption{Workflow of the Knowledge Augmented Clustering (\KNAC) approach.}
\label{fig:workflow}
\end{figure}

In this paper, we present the Knowledge Augmenting Clustering (\KNAC) approach, which can be applied as a form of extension to an arbitrary existing clustering algorithm.
The main goal of \KNAC is to confront the existing expert knowledge about patterns that exist in the data with the results obtained from the automated algorithms and recommend possible modifications of the prior expert knowledge.
In such an approach, the expert knowledge after the modifications is retained and can be used later in the system or during further iterations of clustering.
Figure~\ref{fig:workflow} presents the main functionality of our solution and shows its place in the classic DM/ML pipeline.

There are two parallel processes of labelling the data;  
one is performed by an expert and the other with an automated clustering algorithm.
Initially, the pipeline of the automated algorithm is not affected in any way by the knowledge-based clustering.
However, there is a possible interaction between these two.
After both labellings are finished, the results are delivered to \KNAC which performs cross-checking of the homogeneity and completeness of automated clustering with respect to expert labelling.
Based on that, recommendations of possible splits and merges are generated and presented to the expert, along with appropriate explanations in the form of human-readable rules.
Finally, the expert, equipped with all of this information, can modify the prior knowledge and repeat the process iteratively until a convergence is achieved.
This work is a continuation of the preliminary results published in~\cite{sbk2021iccsknac} but substantially extended.
The extension concerns primarily updated recommendation and justification algorithms as well as a comprehensive evaluation study with an industrial case and publicly available benchmark dataset that demonstrate the feasibility of our approach.

Our work is carried out within the  PACMEL project under the CHIST-ERA program.\footnote{See the project webpage at \url{http://PACMEL.geist.re}.}
The main project goals concern the development of novel methods of knowledge modelling and intelligent data analysis in Industry 4.0.
In this paper, the industrial case study is related to underground mining facilities.
The data used for evaluation was delivered by our industrial partner, Famur S.A.\footnote{See the company webpage at \url{http://www.FAMUR.com.pl}.}, which is one of the global suppliers of coal longwall mining machines used in the so-called longwall mining process.

The rest of the paper is organised as follows:
in Sec.~\ref{sec:sota} we present related works in the area of knowledge-enriched and explainable clustering.
We summarise the state-of-the-art solutions and provide a brief description of our original contribution.
A detailed description of \KNAC is given in Sec.~\ref{sec:knac} along with artificial examples.
A real-life use case scenario is presented in Sec.~\ref{sec:case}.
Then, in Sec.~\ref{sec:eval}, we provide the results of the evaluation with a larger group of experts, based on a publicly available dataset from the e-commerce business domain.
Finally, we summarise our work and discuss the limitations of \KNAC in Sec.~\ref{sec:summary}.

\section{Related works}
\label{sec:sota}
Clustering aims at unfolding hidden patterns in data to discover similar instances and group them under common cluster labels.
This task  is often performed to either discover unknown groups, to automate the process of discovering possibly known groups, or for segmentation of data points into an arbitrary number of segments. 
Any of the above can be done in an unsupervised, semi-supervised, or supervised manner, which depends on the availability of the prior knowledge and the ability to derive new knowledge based on partial data and human interaction.

In this section, we present a landscape of related works that consider both trends in combining clustering algorithms with background knowledge.
The first trend is focused on obtaining  clusters of better quality.
The second trend is concentrated on deriving new knowledge on top of the analysis of the structure of the discovered clusters.
Both of these research areas contributed equally to the motivation of our work, which we present at the end of this section along with a comparison of existing solutions.

\subsection{Knowledge-enriched clustering}
\label{sec:sec:semclust}
In many practical applications, prior knowledge is available for machine learning algorithms to be utilised. 
However, incorporating it into the statistical learning pipeline is a non-trivial task and has been a matter of study for decades~\cite{vonrueden2020informed}.

In~\cite{COLETTA2019150}, the authors combine classification and clusterization to obtain methods for detecting new classes of images that did not appear in the dataset during training.
They transfer knowledge of a previously trained classifier by improving the C3E algorithm~\cite{c3e2011acharyya}, which  works under the assumption that similar instances found by clustering algorithms are more likely to share the same class label obtained with the classifier.
The authors exploited this knowledge and observed that the lack of consensus may indicate a new cluster appearing in the dataset.
Therefore, the candidates for new clusters are those from high-density regions and with high classification entropy (with respect to the known classes).
Such candidates are later labelled by an expert.
However, no prior knowledge is used in this approach, nor are the final suggestions for the expert formulated in a semantically meaningful way, which makes the interpretation of the recommendations highly domain dependent and requires lots of expertise in the domain.


The human-in-the-loop paradigm was also investigated in the clustering algorithms~\cite{contex-clustering}, where the contextual information and user feedback are used to merge clusters of photographs into larger groups.
The background knowledge is partially inferred from the data and encoded in the form of context constraints, that co-occurrence in automatically discovered clusters may impose the need for merging or splitting clusters.
The process is iterative and the human feedback is used as a kind of stop criterion.
However, the process of clusters' refinement is not formalised nor recorded,
making it unsuitable for further use.
Similarly, in other approaches that incorporate domain knowledge into the process of clustering via direct human interaction, the knowledge itself is not retained after the process ends and it cannot be reused later nor checked against existing formalised knowledge~\cite{clustervision2018,interactivetopic2016,interclust2020nowaczyk}.


A different approach was given in~\cite{semiclust}, whereby the authors propose an extension to the k-means clustering algorithm 
that employs constraint information in the definition of multiple assistant representatives
for the centroids used at each iteration of k-means.
These assistant centroids are computed from the data and are considered additional knowledge in forming the final constraints for the clustering algorithm.
Although the external knowledge is gained through the process of generation of assistant centroids, such knowledge is not retained for further use.
Furthermore, the approach does not assume the incorporation of existing domain knowledge into the process of constructing the assistant centroids.


In~\cite{REN2019121}, the authors propose a semi-supervised clustering algorithm that exploits pairwise constraints as background knowledge and integrates it with a deep neural network architecture for clustering.
The integration of background knowledge is done at the level of embedding of feature space in the form of an additional layer.
This influences the embedding and, hence, impacts the clustering results obtained on the latent space.
However, the entire knowledge compilation is done in the latent space and is hidden from the human outside the black-box of DNN.
Therefore, it cannot be easily formalised, nor expressed semantically.


In~\cite{FORESTIER2010211}, the authors present an algorithm for collaborative clustering that aims at finding a consensus between the clustering results from an ensemble of automated methods. 
This method is similar to our approach as one of the artificial methods can be substituted by a human operator.
The method proposes a 3-step approach: initial clustering, results refinement, and consensus computation via a voting algorithm.
However, the input knowledge is not formalised in any way, as it is solely based on automated algorithms and
follows the pattern of introducing a weighted constraints approach that allows incorporating different types of constraints as background knowledge, including  cluster quality, class label, and link-based constraints.
Furthermore, the refinement process of the algorithm is not fully transparent to the user and does not produce any additional knowledge which may indicate how the initial background knowledge should be altered.



In~\cite{YU2020823}, the authors provide an algorithm that uses active learning to obtain constraints utilised later in the clustering process.
These constraints are formed from human responses to selected queries.
The authors focus in their work on highly dimensional cases and, therefore, implement an additional layer that limits the number of queries to the user by selecting dimensions/features that are most informative for the clustering mechanism.
However, the information about these informative features and the way they affect the clustering process is not stored nor formalised for further reuse.
It serves only for the internal algorithm mechanism.


In~\cite{hil-cluster}, the Grouper framework was presented, which is an interactive approval and refinement 
toolkit for the analysis of the results of clustering. 
It combines the strength of algorithmic clustering with the usability of a visual clustering paradigm.
In~\cite{hil-visual-clust}, a similar approach was presented, however, it assumes more interactions with visualised clusters that alters the cluster layout.
Yet, neither of these  uses any kind of formalised knowledge, neither for clustering nor after it for refinement.
Therefore, the knowledge inputted by an expert in the form of interaction in the system is lost for further reuse.
%
Finally, in~\cite{transferclust2017}, the authors present a transfer learning approach that utilises knowledge obtained from similar tasks in the new domain via a fine-tuning procedure.
Transfer learning allows the encoding of domain knowledge into the embedding space that can later be used and fine-tuned to different tasks.
However, the embedding space is latent  and, therefore, cannot be directly utilised, nor confronted with existing domain knowledge.
This limits the usage of the approach in a setting where symbolic knowledge is already present and used in its symbolic form.

\begin{table}[]
\small
\begin{tabularx}{\textwidth}{|X|X|X|X|X|X|}
\hline
{\textbf{Reference}}                        & {\textbf{Background knowledge}}                  & {\textbf{Integration method}} & {\textbf{Integration level}} & {\textbf{Prior knowledge refinement}} & {\textbf{Robustness}} \\ \hline
{\cite{toon2018constraintclust}}   & {constraints}                                    & {human interaction}           & {cluster prototype}          & {no}                                  & {independent}  \\ \hline
{\cite{COLETTA2019150}}    & {ML model}                                       & {human interaction}           & {cluster-prototype}          & {no}                                  & {domain dependent}    \\ \hline
{\cite{semiclust}}         & {statistics derived from data}                   & {constraints satisfaction}    & {cluster-prototype}          & {no}                                  & {independent}  \\ \hline
{\cite{contex-clustering}} & {informal expert knowledge, context constraints} & {human interaction}           & {instance}                           & {no}                                  & {domain-dependent}                    \\ \hline
{\cite{REN2019121}}        & {constraints}                                   & {constraint embedding}        & {cluster-prototype}                           & {no}                                  & {model dependent}                    \\ \hline
{\cite{FORESTIER2010211}}  & {heterogenous constraints}                      & {constraint optimization}     & {cluster-prototype}                           & {no}                                  & {independent}                    \\ \hline
{\cite{YU2020823}}         & {pairwise constraints}                           & {human interaction}           & {instance}                   & {no}                                  & {independent}  \\ \hline
{\cite{transferclust2017}} & {ML model}                                       & {transfer learning}           & {instance}                   & {no}                                  & {model dependant}     \\ \hline
\cite{hil-cluster,hil-visual-clust,interactivetopic2016} & informal expert knowledge                      & human interaction           & cluster                    & no                                  & domain dependent    \\ \hline
\cite{clustervision2018} & informal expert knowledge                      & human interaction           & cluster                    & no                                  & {independent}    \\ \hline            
\end{tabularx}
\caption{Comparison of knowledge augmented clustering methods.}
\label{tab:knac}
\end{table}

\subsection{Explainable clustering}
\label{sec:clex}
Explainable AI (XAI) aims at bringing transparency to the decision-making process of automated systems~\cite{molnar2020interpretable}.
It has been extensively developed over the last decade, mostly due to the rapid development of black-box machine learning algorithms such as deep neural networks.
However, its potential usage is not limited to these methods and is expanded to other artificial AI areas.
In terms of clustering, it is most often used in order to motivate the assignment of a single instance to a particular cluster or to explain the difference between discovered patterns.

In~\cite{explainit2019}, the authors present EXPLAIN-IT: a mechanism that transforms an unsupervised clustering task into a supervised classification task, and use LIME~\cite{lime} and SHAP~\cite{shap} to explain the differences between them.
They demonstrate the usage of their solution on the YouTube video quality classification based on the traffic information.
However, in their approach, the authors do not use any prior knowledge that is later enhanced with their explanation mechanism.
Furthermore, they do not formalise the knowledge discovered with the XAI method for further use.
Alternatively, they limit the explanation only to the visualisation of the contribution of different features to the surrogate classifier built on top of the clustered data.

In~\cite{exclusttext2016}, the authors discuss an explainable clustering algorithm that can be used to group texts, which are represented as a multidimensional dataset. 
They have taken into consideration time-varying changes in a group of texts, hence, bringing the problem closer to time-series clustering. 
The authors emphasize the explainability of their method by providing a highlight for words that are important in assigning an instance to a particular cluster. 
However, no prior knowledge is utilised in this approach.

In~\cite{dasgupta2020explainable}, the authors present a novel clustering algorithm which is inherently explainable. The explanation granularity is performed on the instance level. They use a decision tree structure to obtain both clustering and explanations in the form of rules built from decision tree branches. The decision tree is built in an unsupervised manner, redefining the split criterion in terms of a sum of squared errors with medians or medoids as centroids.

In~\cite{horel2020explainable}, the authors perform a two-step explanation procedure.
First, they obtain cluster labels with an arbitrarily selected clustering mechanism, which later are used as the target variable in a classification task.
The classification is finally explained using a Single Feature Introduction Test (SFIT)~\cite{horel2019computationally} to denote statistically important features that take part in the classification process.
However, the obtained explanations are feature importances that do not provide enough expressive power to be considered stand-alone explanations, as additional domain or expert knowledge is required to understand the way the features affect the classifier.

In~\cite{frost2020exkmc}, the authors present a similar solution to that discussed in~\cite{dasgupta2020explainable} by exploiting decision trees as clustering mechanisms.
They emphasise the issues related to the trade-off between explainability and accuracy in the case of inherently interpretable models.
They introduce the ExKMC algorithm, which is an explainable version of the k-means clustering algorithm.
They prove that the surrogate cost is non-increasing and, hence, the aforementioned trade-off is under control.
Although the explanation is given in the form of a rule, obtained from the translation of the tree branches, no prior knowledge is exploited in the process of clustering.


In~\cite{eUD3clust}, the authors explore decision trees to obtain cluster labels, however, in their approach, they build several decision trees and combine the results into the final clustering.
In contrast to previously discussed methods, this approach takes different factors into consideration in forming clusters, i.e., performing splits in a decision tree.
They put the emphasis on the compactness and separation of the cluster as a forming criterion~\cite{JAIN2010651}.
Similar approaches were given in works describing CLUS~\cite{clus} and classic COBWEB~\cite{cobweb} algorithms.



Other methods of exploiting ensemble classifiers were discussed in~\cite{shi2006rfclust,Kruber2018rfclust,madhyastha2019geodesic}, where the random forest algorithm was chosen.
The explanation is then obtained  on the global level by generating feature importance computed as the (normalized) total reduction of the criterion brought by that feature. 
Alternatively, model agnostic explanation algorithms such as LIME or SHAP are also applied to generate local (instance-level) explanations~\cite{rodrig2015rfclust}.

In~\cite{excut2020}, the authors introduce ExCut, an approach for computing explainable clusters, which combines embedding-based clustering with symbolic rule learning to produce human-understandable explanations for the resulting clusters.
The method is designed for knowledge graphs, and its goal is to cluster semantically similar entities as denoted by the underlying background knowledge component.
The huge advantage of this method is the simplicity in introducing background and/or expert knowledge into the process, as the input dataset is itself a knowledge graph, which may contain expert knowledge.
The discovered knowledge can later be incorporated into the knowledge graph.
This approach is similar to our contribution, except it makes the explanations and knowledge enhancements an internal component of clustering.
This exposes some limitations in applications of this method on a meta-level, where the clustering methodology should be considered independent from the explanation and expert-knowledge encoding formalism.


Similar approaches that utilize knowledge graphs for providing an explanation of clustering were discussed in~\cite{dedalo2014} and~\cite{dedalo2015}, where knowledge graphs are exploited to obtain explanations of previously discovered clusters.
The method in~\cite{kgembedclust} uses knowledge graph embedding to find a pattern which makes similar concepts be placed close to each other in latent space.
Similar solutions exploiting latent space as a search field for combining external and input knowledge were also discussed in~\cite{bouraoui2018learning,hamilton2019embedding}.

In~\cite{xaiclust2021}, the authors put emphasis on delivering explanations of clustering results with a symbolic knowledge representation mechanism.
In order to do this, a methodology based on a surrogate model is presented that, similarly to other approaches, translates the problem of clustering into a classification task with clustering labels as the target variable.
The authors aim to provide rules for the explanation of clustering, however, they do not give any method that allows integrating these rules with prior knowledge.
Furthermore, they mostly use tree-based surrogate models as explanation mechanisms due to the limitations of the frameworks they use.
Finally, the approach presented in their paper focuses on explanations of single instances belonging to particular clusters, not a summary that can be valid for the whole cluster.
This results in overwhelmingly large explanations for large datasets, which might not be useful for the end user.


In~\cite{timecluster}, the authors focus on solutions which assist users to understand long time-series data by observing its changes over time, finding repeated patterns, detecting outliers, and effectively labelling data instances. It is performed mostly via a visualisation layer over the data which dimensionality was reduced with UMAP~\cite{mcinnes2020umap} allowing 2D/3D plotting.
However, no explicit knowledge is used in this method to enhance the process of cluster analysis.

We summarize these relevant results in Tab.~\ref{tab:clex}.

\begin{table}[]
\small
\begin{tabularx}{\textwidth}{|X|X|X|X|X|}
\hline
\textbf{Reference} & \textbf{Explanation form} & \textbf{Explanation mechanism} & \textbf{Prior knowledge} & \textbf{Explanation granularity} \\ \hline
\cite{explainit2019}                                       & statistical summary, visualization & model-agnostic                 & no                       & single instance                  \\ \hline
\cite{exclusttext2016}                                     & example-based                      & model-specific                 & no                       & single instance                  \\ \hline
\cite{dasgupta2020explainable}                             & rules,trees                        & model-specific                  & no                       & single instance                  \\ \hline
\cite{horel2020explainable}                                & feature importance                 & model-agnostic                 & no                       & global                           \\ \hline
\cite{frost2020exkmc}                                      & rules                              & model-specific                 & no                       & single instance                  \\ \hline
\cite{eUD3clust}\cite{JAIN2010651},\cite{clus}\cite{cobweb} & rules, trees                       & model-specific                 & no                       & single-instance                  \\ \hline
\cite{excut2020}                                           & rules, ontologies                  & model-specific                 & yes                      & single instance                  \\ \hline
\cite{xaiclust2021}                                        & rules                              & model-agnostic                 & no                       & single instance                  \\ \hline
\cite{Loyola-Gonzalez2020}                                        & rules                              & model-agnostic                 & no                       & global                  \\ \hline
\cite{kgembedclust}                                        & knowledge graph                                & model-specific                 & no                       & global                  \\ \hline

\end{tabularx}
\caption{Comparison of explainable clustering methods.}
\label{tab:clex}
\end{table}

\subsection{Our original contribution}
\label{sec:orig}
Although there exists a large spectrum of clustering methods that are able to incorporate external sources of knowledge to enhance the process of clustering or provide explanations for the clustering results they have calculated, apparently there is no research in the area of combining these two into one comprehensive framework.
We argue that in many practical applications, where prior knowledge is available, it is not only important to incorporate this knowledge into the clustering pipeline but also refine it with the results obtained from the clustering algorithms.
In many cases, the prior knowledge is coarse and too general to capture some patterns that are visible in  the real-life scenarios.
As shown in the comparison presented in Tab.~\ref{tab:knac}, clustering methods that use prior knowledge most often are focused on boosting the clustering performance and quality. 
No further refinement of the initial knowledge is assumed. 
There are several methods that use knowledge graphs as a source of knowledge and allow online modifications of the prior knowledge source.
However, these refinement methods are tightly connected with the clustering algorithm itself, making them lack robustness and not be easily extensible to other clustering approaches.
In~\cite{interclust2020nowaczyk}, the authors present a comprehensive survey on interactive clustering, comparing and deeply analysing over 100 papers from the field.
They notice the same gap in the methods that we did, which is the lack of an algorithmic approach that will support cluster analysis, observing that:
\begin{quote}
There is a need for developing solutions where the machine would initiate quality improving operations, for example, indicate to the users specific clusters that require feedback in a form of a query, and the users would provide information based on such requests. 
\end{quote}
We additionally claim that such machine-initiated operations should be \emph{explainable and persistent}, even after the clustering algorithm is finished for post-hoc analysis of the clustering or expert knowledge refinement.
However, the explainable clustering algorithms summarised in Tab.~\ref{tab:clex} do not make any use of prior knowledge or are tightly bounded with specific clustering algorithms, making the whole method barely applicable to other cases.

Taking all of the above into consideration, we defined three requirements that the method should follow to solve the problem of combining clustering explanations with the usage of prior knowledge and refinement of this knowledge.
We argue that such a method should be: 
\begin{itemize}

\item (R1): Human readable -- the prior knowledge, its refinement should be delivered in a human-readable form. This requirement is mostly affected by the explanation form and granularity of explanation. 
\item (R2): Executable -- it should be possible to automatically process the prior knowledge and its refinements and integrate it with state-of-the-art tools supporting the DM/ML pipeline. This requirement depends on the explanation mechanism used and the prior knowledge encoding method. 
\item (R3): Approach agnostic -- it should be possible to apply these methods independently of the clustering algorithm and the clustering pipeline that was chosen. This requirement is mostly affected by the explanation mechanism and integration capabilities of the prior knowledge encoding mechanism. 
\end{itemize}

Therefore, we propose a method that uses rule-based notation for encoding prior knowledge and its refinements.
This is one of the most human-readable methods of encoding knowledge that can at the same time be executable (R1).
Rules in their preconditions use the same set of features that an expert operates on, although the automated clustering pipeline can include an unlimited number of transformations of an initial dataset.
We use the \XTT notation for encoding rules and  a \HRTD inference engine to execute them (R2) and integrate them with external system components~\cite{sbk2018exsy}.
Our approach can be considered an augmentation method that can be applied to any selected clustering algorithm, as the knowledge refinement is performed independently on the internal implementation of the clustering algorithms (R3).

As mentioned in the introduction, this work is a continuation of the preliminary results published in~\cite{sbk2021iccsknac}; however, substantially extended.
Specifically, in the work presented here, we changed the algorithm which is responsible for conformance checking between domain knowledge and automatically discovered clusters, and for the generation of split and merge recommendations.
Previously, it depended only on the contingency matrix and simple statistics calculated based on it.
Here, we allow measuring the quality of the refinements and adjusting the split and merge recommendations taking those into consideration.
Furthermore, in this work, we conducted experiments on two cases with more than 30 participants involved in total.
One of the experiments, along with its dataset, is publicly available for benchmark and comparison purposes.

In the following sections, the details of the underlying mechanisms of our method are presented.

\section{The knowledge augmented clustering approach}
\label{sec:knac}
The main goal of the work presented in this paper is to refine the initial clustering with XAI methods and expert knowledge via splitting and merging the clusters delivered by an expert 
and generating explanations (or justifications) of these recommendations.
The goal of these refinements is to help the experts to better understand the patterns that exist in the data and, therefore, obtain clusters which are of better quality.
In this section, we present the theoretical background of our approach and demonstrate its feasibility on synthetic examples to make the work reproducible and transparent\footnote{For source code with reproducible examples see: \url{http://github.com/sbobek/knac}}.
All presented datasets were generated using two functions from scikit-learn library with a \verb|make_blobs| function which generates isotropic Gaussian blobs.
Every synthetic dataset presented in this paper is 2-dimensional for the sake of simplicity of presentation.
However, we provided more examples in a publicly available repository, including 
a multidimensional case with  20 features\footnote{See: \url{http://gitlab.com/sbobek/knac}}.

Below, we assume the following.
There exist a set of clusters obtained with expert knowledge, denoted as: $$E=\left \{ E_1, E_2, \ldots, E_n \right \}$$
This set of clusters needs to be refined with complementary clustering performed with automated clustering algorithms. This clustering forms a separate set of clusters, possibly of different sizes than $E$ and is denoted as: $$C=\left \{ C_1, C_2, \ldots, C_m \right \}$$

For both sets, we calculate a contingency matrix $M$ of size $(n,m)$ where the number at the intersection of the $i$-th row and $k$-th column holds the number of data points assigned both by cluster labelling to cluster $E_j$ and automated clustering to cluster $C_k$.

Based on the contingency matrix $M$, we calculate two helper matrices for splitting and merging strategies defined respectively by Eq.~(\ref{eq:mat:split}) and~(\ref{eq:mat:merge}). 
In the case of $H^{split}$, we additionally $l2$ normalize the $M$ matrix along columns to appropriately deal with clusters of different sizes.

\begin{equation}
    H_{i,j}^{split} =  \frac{M_{i,j}}{||M_i||_2 \left [\frac{ H(M_i)}{log2(||E||)}+1 \right ]}
    \label{eq:mat:split}
\end{equation}

Where $H(M_j)$ is entropy calculated for the $j$-th column (i.e., $C_j$ cluster). 
The measure defines the consistency of automated clustering with expert clustering.
We perform  min-max scaling along the rows of $H^{split}$ to allow thresholding from a range $[0;1]$.
Because the maximum value of $H(M_j)$ is an entropy measure and depends on the number of expert labels, we normalise the $H^{split}$ matrix accordingly.

For the merging operation, we calculate the $H^{merge}$  matrix.
It is a $l2$ normalised matrix $M$ along the row axis.

\begin{equation}
    H_{i,j}^{merge} = \frac{M_{i,j}}{||M_i||_2}
    \label{eq:mat:merge}
\end{equation}

\begin{figure}
\centering
\includegraphics[width=\textwidth]{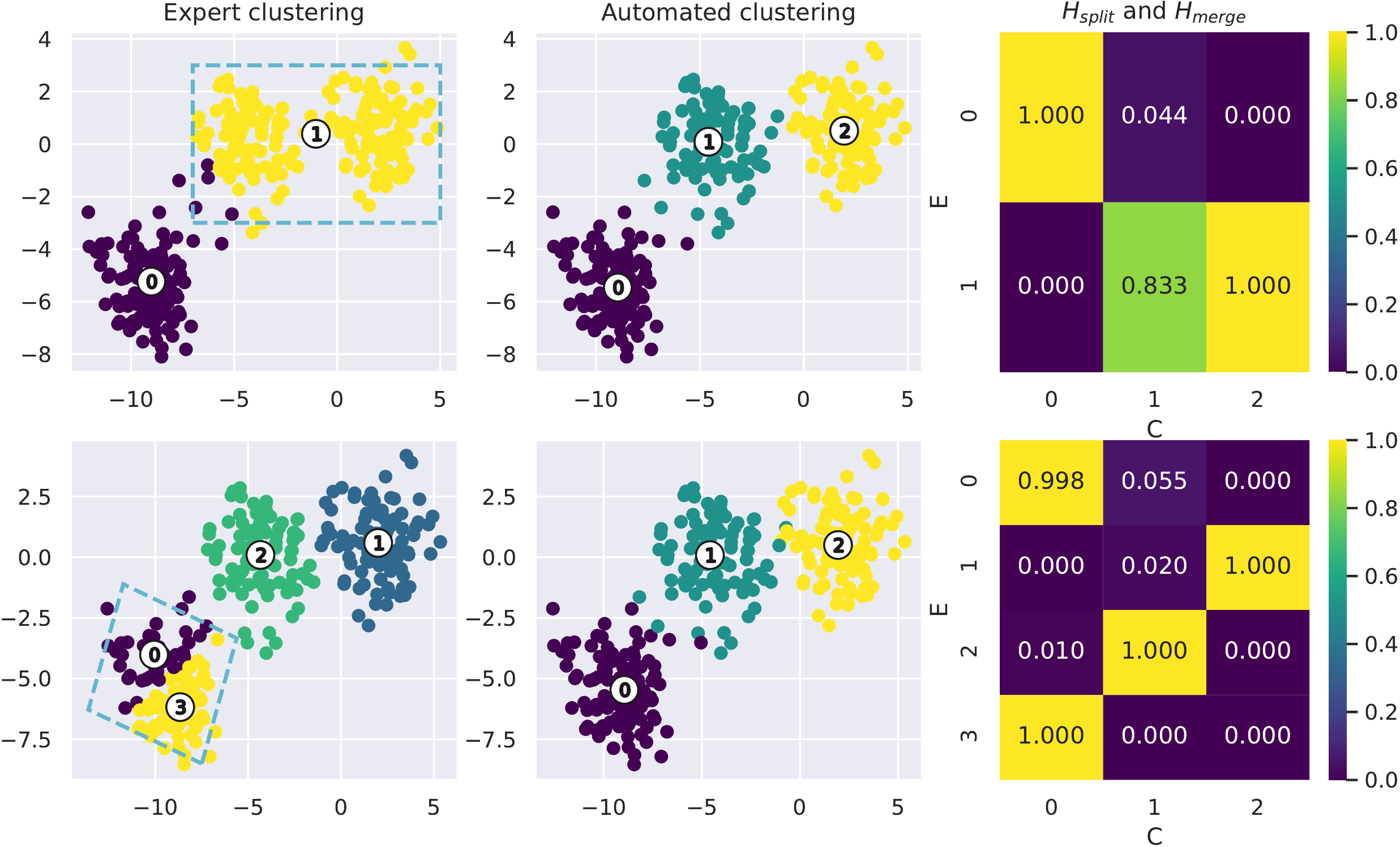}
\caption{Synthetic datasets with clusters to split (top row) and clusters to merge (bottom row). Columns in the figure represent clustering performed with expert knowledge, automated clustering, and $H^{split}$ matrix (upper) and $H^{merge}$ matrix (lower). Dotted lines define bounding boxes for the decision stump explanation mechanism.}
\label{fig:synthds}
\end{figure}

These two matrices are later used for the purpose of generating split and merge recommendations.
Fig.~\ref{fig:synthds} depicts the two simplified datasets with two possible scenarios covered by our method. 
These datasets will be used to better explain the mechanisms for the generation of splitting and merging recommendations discussed in the following sections.

\subsection{Recommendation generation}
\label{sec:split_merge_recoms}
Having created the $H^{split}$ and $H^{merge}$ matrices, we generate two types of recommendations out of it: \emph{splitting} and \emph{merging} of initial expert clustering.

\paragraph{Splitting.} This recommendation aims at discovering clusters that were incorrectly labelled using expert knowledge.
Such a case was depicted in Fig.~\ref{fig:synthds} in the upper left plot.
This operation can be performed using the $H^{split}$ matrix in a straightforward way.
The cluster that is recommended for splitting is chosen by investigating values corresponding to columns of a selected row in the $H^{split}$ matrix.
Values that lie on the intersection of the investigated expert cluster and the automated cluster and are greater than the defined confidence threshold $\epsilon_s$ are marked as candidates for splitting.

Because the pure $H^{split}$ matrix is based only on the distribution of the points between expert labelling and automated labelling, we additionally recalculate confidences for particular splits by checking the relative decrease in the silhouette coefficient $S^{dec}$ for every potential split.
Due to the large computational overhead of calculating the silhouette coefficient, we only calculate it for the fraction of candidates.
The proportion of additional candidates taken into consideration while calculating $S^{dec}$ is balanced by the silhouette weight  parameter $\lambda^{s}$ in Eq.~(\ref{eq:split:cands}).
The larger the $\lambda^{s}$, the more candidates are taken into consideration while calculating $S^{dec}$.

\begin{equation}
C^{split}_i=\left \{c_j \in  H^{split}_{i}: \frac{ c_j  }{1-\lambda^{s}} > \epsilon_s \right \}
\label{eq:split:cands}
\end{equation}

Once the candidates are selected, the confidence for each of them to be included in the final split is calculated.
The relative decrease in the silhouette coefficient $S^{dec}(C^{split}_i)$ is calculated as a difference in the silhouette coefficient calculated for the dataset before and after the potential split with the set defined by $C^{split}_i$, scaled to the range of $[0;1]$.
The trade-off between the confidence derived from $H^{split}$ and the one based on the relative silhouette coefficient decrease is also governed by the $\lambda^{s}$ parameter.
The confidence for each of the split candidates $Conf(C^{split}_i)$ is calculated according to Eq.~(\ref{eq:split:confs}).
The final confidence of the rule defining a split is calculated as an average of the confidences of the candidates that form the rule:

\begin{equation}
\mathit{Conf}(C^{split}_i) = \left \{ (1-\lambda^{s})c_j + \lambda^{s}(S^{dec}(C^{split}_i)) : c_j \in C^{split}_i \right \}
\label{eq:split:confs}
\end{equation}

For the example in Fig.~\ref{fig:synthds}, the recommendation will look as follows, with the silhouette coefficient weight $\lambda^{s}=0.1$ 

{\small
\begin{verbatim}
SPLIT 
    EXPERT CLUSTER  E_1 
INTO 
    CLUSTERS  [(C_1, C_2)]  (Confidence 0.87)
\end{verbatim}
}


\paragraph{Merging.} The goal of this recommendation is to detect concepts that were incorrectly labelled by expert knowledge as two or more clusters. 
Such a case is depicted in Fig.~\ref{fig:synthds} in the lower left plot.
Candidates for merging are chosen using the $H^{merge}$ matrix.
Because the matrix is $l2$ normalised along the rows, calculating a dot product of selected rows produces cosine similarity between them.
This reflects the similarity in the distribution of data points spread over the automatically discovered clusters.
If two expert clusters have a similar distribution of points over automatically discovered clusters, this \emph{might} be a premise that they share the same concept and should be merged.
Such a case was depicted in Fig.~\ref{fig:synthds} in the lower right plot. 
Similar to the case of splitting, a threshold $\epsilon_m$ is defined arbitrarily denoting the lower bound on the cosine similarity between clusters to be considered as merge candidates.

In the same way as in the case of splitting, we allow for updating the confidence of merges by  exploiting linkage strategy from hierarchical clustering.
For every expert cluster, we calculate a distance matrix $D^{linkage}$ with one of the following methods: single-link, complete-link, centroid-link, and average link.
We also calculate a similarity matrix $H^{sim} = H^{merge} \cdot (H^{merge})^\intercal$, where each of the cells $j,k$ represents the cosine similarity between the distribution of data samples of expert cluster $j$ and $k$ over clusters $C$ discovered with the automated method.
We then combine the similarity matrix $H^{sim}$ and distance matrix $D^{linkage}$ to calculate the final confidence value $C^{merge}_{j,k}$ for merging expert clusters $j$ and $k$, as defined in Eq.~({\ref{eq:merge:candidates}).

\begin{equation}
C^{merge}_{j,k} = \left \{E \ni E_j, E_k: (1-\lambda^{m})H^{sim}_{j,k}+\lambda^{m}(1-D^{linkage}_{j,k}) > \epsilon_m 
\right \}
\label{eq:merge:candidates}
\end{equation}

The merge recommendation for the case depicted in Fig.~\ref{fig:synthds} is given below with the linkage metric weight $\lambda^{m}=0.2$:

{\small
\begin{verbatim}
MERGE 
    EXPERT CLUSTER E_0 
WITH 
    EXPERT CLUSTER E_3 
INTO 
    CLUSTER C_0 # (Confidence 0.98)
\end{verbatim}
}

The confidence value is calculated as the weighted sum of  a cosine similarity between rows associated to candidates $E_0, E_3$ in the $H^{merge}$ matrix and the distance between these two clusters according to the selected linkage strategy.
In the next section, the justification mechanism of the split and merge recommendation is discussed.

\subsection{Explanation of recommendation}
Once the recommendation is generated, it is augmented with an explanation.
Depending on the recommendation type, the explanation is created differently.

\paragraph{Splitting recommendation.}
In the case of this type of recommendation, we transform the original task from clustering to classification, taking the automatically discovered cluster labels as target values for the classifier.

Then, we explain the decision of a classifier to present to the expert why and how the two (or more) clusters $C_i, C_j, \ldots,C_n$ that were formed by splitting the original one are distinguished from each other.
An explanation is formulated in the form of a rule that uses the original features as conditional attributes, to help the expert better understand the difference between splitting candidates.
We use the Anchor algorithm for this~\cite{anchor}, which is model-agnostic and allows for the explanation of an arbitrarily selected model.
Alternatively, one can use our LUX~\cite{sbk2021iccslux} algorithm, which provides similar functionality but has  better time complexity.
In both cases, the output from the explanation mechanism is translated into \XTT rules, which are a human-readable and executable format of rule-based knowledge.

The explanation for the splitting of cluster $E_2$ presented in Fig.~\ref{fig:synthds} looks as follows:

{\small
\begin{verbatim}
C_1: x1 <= -8.20 AND x2 > -4.34 (Precision: 1.00, Coverage: 0.07)
C_2: x2 <= -4.34 (Precision: 0.90, Coverage: 0.25)
\end{verbatim}
}

If the difference is important, the clusters can be split and the rules generated above can be added to the knowledge base.
The final decision on whether splitting $E_2$ into $C_1$ and $C_2$ is needed is left to the expert.

The rules presented in the listing above cannot be executed. 
We, therefore, translate them to the \XTT format which is executable by the \HRTD inference engine~\cite{sbk2018exsy}.
The precision and coverage delivered by Anchor are translated to the rule confidence used by \HRTD as a product of the two former factors.
In the \XTT format, rules can be grouped into tables.
The table for the rules above is presented in Fig.~\ref{fig:artif:splits}.
If the expert decides that the split is relevant, it is enough to include the \XTT rules in the knowledge base, optionally adjusting their confidences.
The \HRTD inference engine will process all the rules (old and new) and decide which cluster assignment is the most certain.

\begin{figure}
\centering
\includegraphics[width=0.8\linewidth]{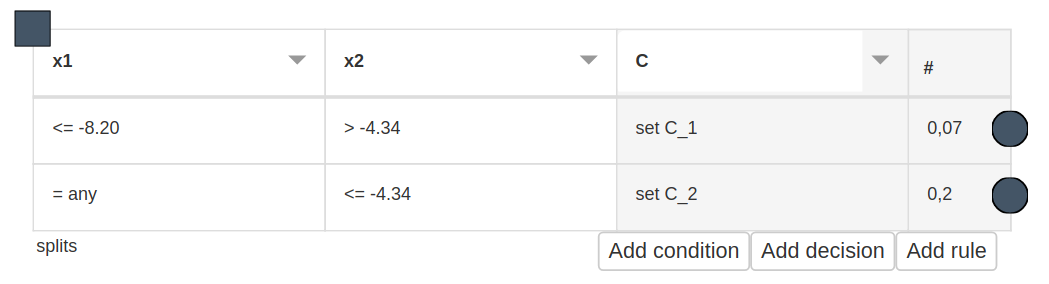}
\caption{\XTT table for rules explaining split recommendation.}
\label{fig:artif:splits}
\end{figure}



%

\paragraph{Merging recommendation.}
In explaining  the merging recommendation, we use the same approach as previously described.
The difference is that the classification models are now trained with expert labels as a target.

After this, the explanation that answers the question as to how the expert clusters $E_i$ and $E_j$ are different from each other is generated by looking at them not through the definition in the knowledge base, but through the data perspective.
The answer to this question is given in the form of Anchor rules, and the final decision is left to the expert.
The explanation for the case presented in Fig.~\ref{fig:synthds} is given below:

{\small
\begin{verbatim}
E_0: x2 <= 5.16 AND x1 > 0.23 (Precision: 0.74, Coverage: 0.32)
E_1: x1 <= 3.73 (Precision: 0.58, Coverage: 0.50)
\end{verbatim}
}

The \XTT table for the rules above is presented in Fig.~\ref{fig:artif:merges}.
If the expert decides that the differences between two expert clusters are irrelevant with respect to the distinction made by the Anchor rules, he or she may decide to merge those clusters.
Otherwise, clusters and prior knowledge remain unchanged.
In contrast to split recommendations, merges require more delicate knowledge base modifications, which are left solely to the expert.
The Anchor rules presented above may only serve as hints on how to modify the knowledge base (e.g., which parameters are irrelevant).

\begin{figure}
\centering
\includegraphics[width=0.8\linewidth]{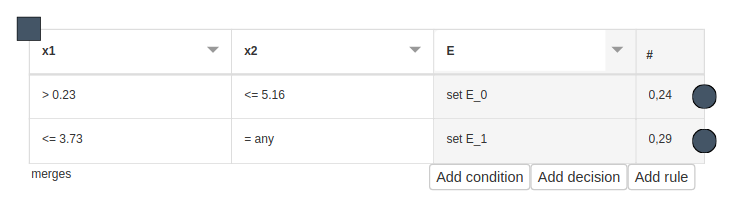}
\caption{\XTT table for rules explaining merge recommendation.}
\label{fig:artif:merges}
\end{figure}



\subsection{The \KNAC algorithm}
In this section, we show \KNAC in the form of a concise procedure presented in Algorithm~\ref{alg:knac} that summarises all of the steps illustrated in previous sections.
The input for the algorithm consists of the dataset $D$, the knowledge base $KB$, and the clustering pipeline for dataset $D$.
The knowledge base $KB$ is considered to be rule-based, however, in  general cases, it can also be a set of labels for each of the instances in $D$.
The algorithm starts by creating a contingency matrix from labels obtained from $KB$ and $CLA$.
This contingency matrix is then used to build $H^{split}$ and $H^{merge}$ matrices as described in Sect.~\ref{sec:split_merge_recoms}.
Based on these structures, the split and merge candidates are created, along with their justifications.
The justifications are assumed to be in the form of rules, however, in general cases, they can be any type of explainable structure that allows for labelling data and explaining differences between particular clusters.
If the expert decides that the justification is convincing, a decision on including a new rule in $KB_{+}$ can be made.
The algorithm converges when no more modifications to $KB_{+}$ are performed during the last pass.

It is worth noting that the elements of $KB_{+}$ do not have to be rules. 
In particular, they can take the form of an instance of clustering algorithms followed by accepted \KNAC recommendations and their explanations (not necessarily in the form of rules).
Such justification can serve as a single interpretable clustering step that makes the whole process more transparent.

\RestyleAlgo{ruled}
\begin{algorithm}[hbt!]
\caption{KnAC algorithm}\label{alg:knac}
\KwData{$D$ -- dataset, $KB$ -- initial domain knowledge, $CLA$ -- clustering pipeline}
\KwResult{$D_L$ -- labelled dataset, $KB_{+}$ -- refined domain knowledge}

$KB_{+} \gets KB$\;
$refine \gets true$\;
\While{$refine \neq true$}{
    $refine \gets false$\;
    $E \gets expert\_label(D,KB_{+}) $\;
    $C \gets automatic\_cluster\_label(D, CLA)$\;
    $M \gets contingency\_matrix(E,C,DS)$\;
    $H_{i,j}^{split} =  \frac{M_{i,j}}{||M_i||_2 \left [\frac{ H(M_i)}{log2(||E||)}+1 \right ]}$\;
    $C^{split}_i=\left \{c_j \in  H^{split}_{i}: \frac{ c_j  }{1-\lambda^{s}} > \epsilon_s \right \}$\;
    $SplitRule_{i} \gets justify(C^{split}_i)$\;

  \If{$SplitRule_{i}$ is accepted by expert }{
    $KB_{+} \gets KB_{+} + {SplitRule_{i}}$\;
    $refine \gets true$\;
  }
  
   $H_{i,j}^{merge} = \frac{M_{i,j}}{||M_i||_2}$\;
    $H^{sim} = H^{merge} \cdot (H^{merge})^\intercal$\;
    $C^{merge}_{j,k} = \left \{E \ni E_j, E_k: (1-\lambda^{m})H^{sim}_{j,k}+\lambda^{m}(1-D^{linkage}_{j,k}) > \epsilon_m \right \}$\;
    $MergeRule_{i,k} \gets justify(C^{merge}_{j,k})$\;
  \If{$MergeRule_{i,k}$ is accepted by expert }{
    $KB_{+} \gets KB_{+} + {MergeRule_{i,k}}$\;
    $refine \gets true$\;
  }
  $D_L \gets expert\_label(D,KB_{+}) $\;
  
}

\end{algorithm}

In the following section, our approach will be presented in a real-life case study scenario from the Industry 4.0 domain.

\section{Case study}
\label{sec:case}

Our industrial use case concerns the operation of a coal mine shearer~\cite{Singh:coal:mining:2014} in an underground coal mine. 
A shearer is the main element of longwall equipment and is used for coal mining and loading on the armoured face conveyor (AFC). A shearer consists of two mining heads (cutter drums), placed on the arms, and a machine body containing electric haulage, hydraulic equipment, and controls. A shearer is mounted over the AFC. The working shearer moves in two directions: along the longwall face (from the maingate to the tailgate), cutting the coal and due to mining direction~-- along the length of the longwall panel. 
From the perspective of this case study, the main challenge is the discovery and identification of the states of the machine, based on the sensor reading monitoring its operation.
To evaluate the automated state detection from data, we used expert knowledge rules to detect ground true labels.
The rules were delivered by a domain expert working in collaboration with the Famur company that delivered the data.

\subsection{Knowledge-based labelling}

\begin{figure}
\centering
\includegraphics[width=0.8\linewidth]{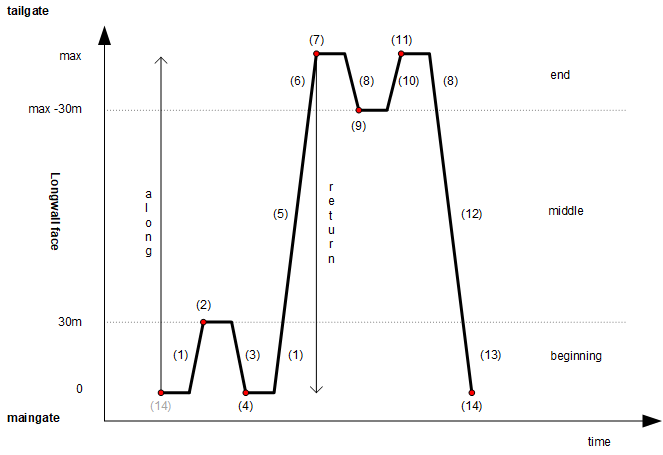}
\caption{Model of a shearer cycle~\cite{msz2020energies}.}
\label{fig:idealcycle}
\vspace{-5mm}
\end{figure}

The rules listed below describe the higher-level operational state of the machine that can be referred to the 
process of coal extraction presented in Fig.~\ref{fig:idealcycle}.
The meaning of the identified process stages in Fig.~\ref{fig:idealcycle} is as follows: 
\begin{enumerate}
\item A - cutting into tailgate direction at the beginning of the longwall,
\item A - stoppage in ON mode at the beginning of the longwall (location: 30-40m from the maingate),
\item A - cutting - return to maingate at the beginning of the longwall,
\item A - stoppage in ON mode at the beginning of the longwall (location: minimal value - maingate),
\item A - cutting in the middle of the longwall,
\item A - cutting into tailgate direction at the end of the longwall,
\item A - stoppage in ON mode at the end of the longwall (maximal value - tailgate),
\item R - cutting into maingate direction at the end of the longwall,
\item R - stoppage in ON mode at the end of the longwall (location: 30-40m from the tailgate),
\item R - cutting - return to tailgate at the end of the longwall,
\item R - stoppage in ON mode at the end of the longwall (maximal value - tailgate),
\item R - cutting in the middle of the longwall,
\item R - cutting into maingate direction at the beginning of the longwall,
\item R - stoppage in ON mode at the beginning of the longwall (location: minimal value - maingate).
\end{enumerate}

The decision tree structuring the expert rules for each state of the shearer, mentioned above (denoted in the picture by an integer value from a range of 1 to 14), is presented in Fig.~\ref{fig:tree}.

\begin{figure}[htb]
\centering
\includegraphics[width=0.67\textwidth, angle=270]{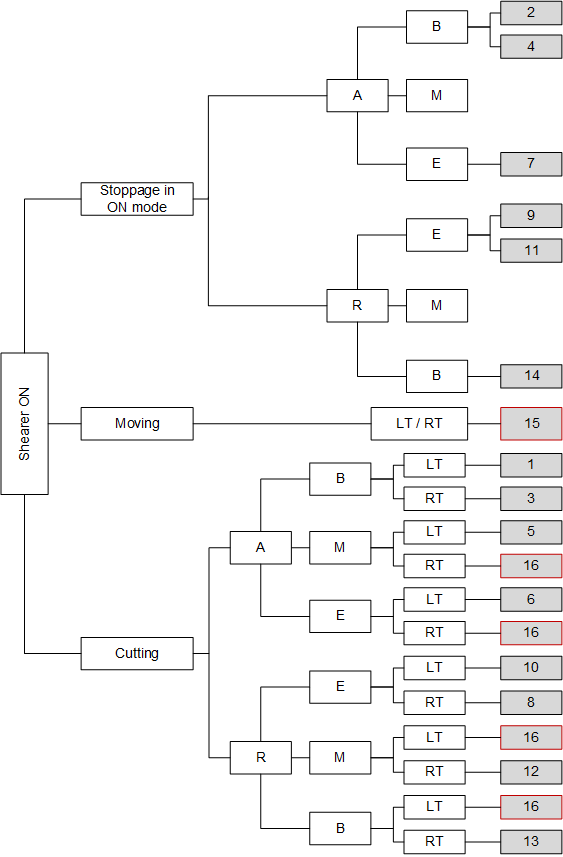}

\caption{Expert tree for theoretical activity description~\cite{sbk2020kr4l}}
\label{fig:tree}
\end{figure}

The first split of the decision tree corresponds to the state of the shearer. There are three main operating states of the shearer: cutting (moving along the longwall face with working drums), moving (moving along the longwall face without working drums), and stoppage.
Specific rules that denote these states are based on currents (haulages and drum) and shearer speed\footnote{Due to the information policy of collaborating companies, the rules cannot be presented in detail.}.
The second split for \emph{Cutting} and \emph{Stoppage in ON mode} is done by part of the cycle: along (A) or return (R). 
The next split depends on the shearer location in the longwall: the beginning (B), the middle (M), or the end (E). 
The last split corresponds to the movement direction of the shearer right (RT) or left (LT). 
\emph{Moving} activity is considered along with a move direction, independently of the location of the shearer. 
In the case of stoppages, the movement direction split is not applicable. 

The presented tree can be expanded into a set of rules, in the form presented below, e.g.:
{\small
\begin{verbatim}
IF shearer state = "cutting"
  AND cycle part = "A"
  AND location = "middle" 
  AND move direction = "LT" 
THEN activity = "5"
\end{verbatim}
}

As can be observed, the decision tree contains two states more than in the original process from Fig.~\ref{fig:idealcycle}, namely moving without cutting (15) and cutting in the opposite direction to the cycle part (16). These states do not directly refer to the regular process of coal extraction.
However, they exist in mining practice, and thus are needed to be considered during process modelling and analysis.

The decision tree for defining the theoretical machinery states depicted in Fig.~\ref{fig:tree} is general enough to cover all possible characteristics of machinery operation in the longwall face.
However, it has to be adjusted in order to fit the available data.
In the real-life scenario presented in this work, the rules were simpler due to the fact that the shearer was cutting only along one direction.
The rule-based representation of the theoretical states is given in the \XTT model presented in Fig.~\ref{fig:rules1}. 
In the figure, we presented only a fragment of the rule-based encoding of theoretical states indicated in Fig.~\ref{fig:tree}.
The rest of the tree is omitted, as it does not play an explicit part in the workflow.

\begin{figure}[htb]
\centering
\includegraphics[width=1\textwidth, angle=0]{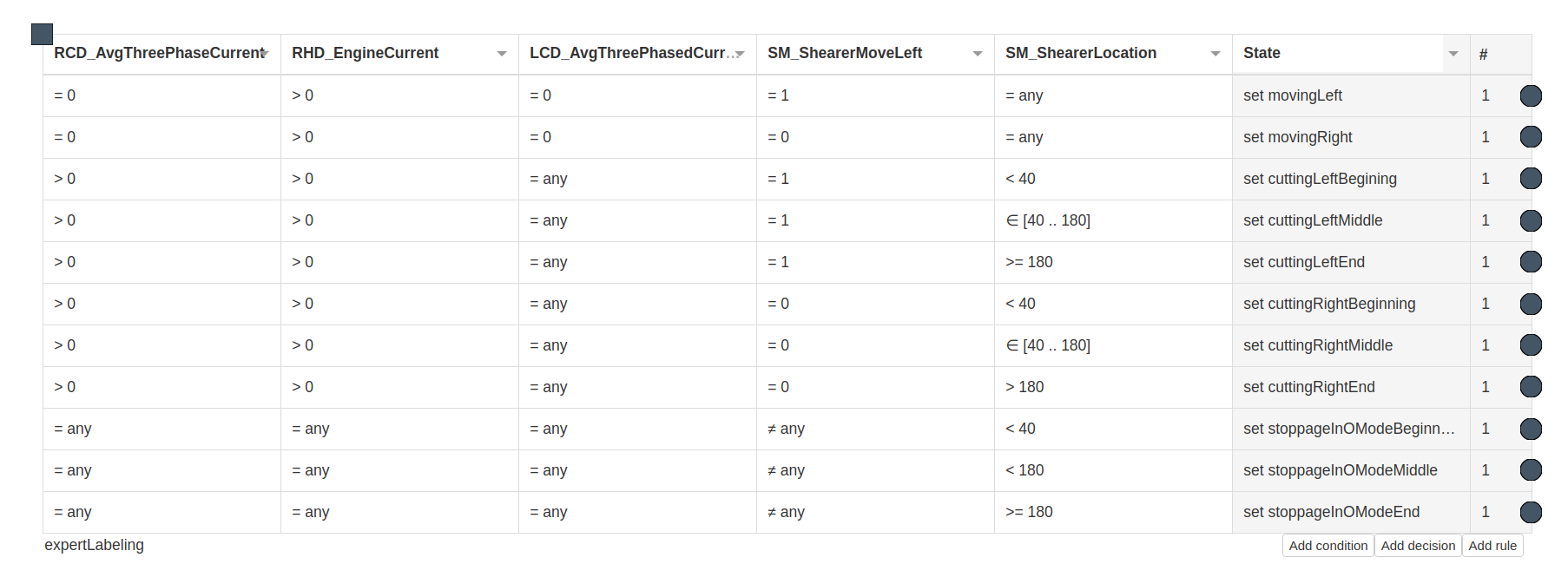}
\caption{Expert rules for theoretical activity description from Fig.~\ref{fig:tree} for defining intermediate clusters of the shearer.}
\label{fig:rules1}
\end{figure}


Rules can be applied directly to label the raw data as they use features that are present in the raw data. 
Such an application of the \XTT rules presented in Fig.~\ref{fig:rules1} is given in Fig.~\ref{fig:expert:stages} and this labelling can extend actual data analysis in the direction of efficiency analysis by generating summaries and basic statistics on the operation of the shearer.
However, in such an approach, some non-typical behaviour of the shearer can be lost.
This may occur when one of the states denoted by the expert rule encapsulates more specific and highly distinguishable states, which can be defined by raw data analysis.
These states may correspond, for example, to abnormal machinery operation due to a possible hardware fault or inappropriate device control by its operator. 

Fig.~\ref{fig:expert:stages} depicts the results of expert label assignment over a selected period of time with the knowledge base shown in Fig.~\ref{fig:rules1}.
The figure depicts the location of the shearer, with colours representing different machinery states (clusters).
In the ideal situation, the expert-based clustering will be perfectly aligned with the theoretical stages depicted in Fig.~\ref{fig:idealcycle}.
However, this is clearly not the case in this example and, hence, additional refinement of the expert knowledge is needed.
Therefore, the expert labelling will be an input for \KNAC along with automated clustering described in more detail in the next section.

\begin{figure}[htb]
\centering
\includegraphics[width=\textwidth]{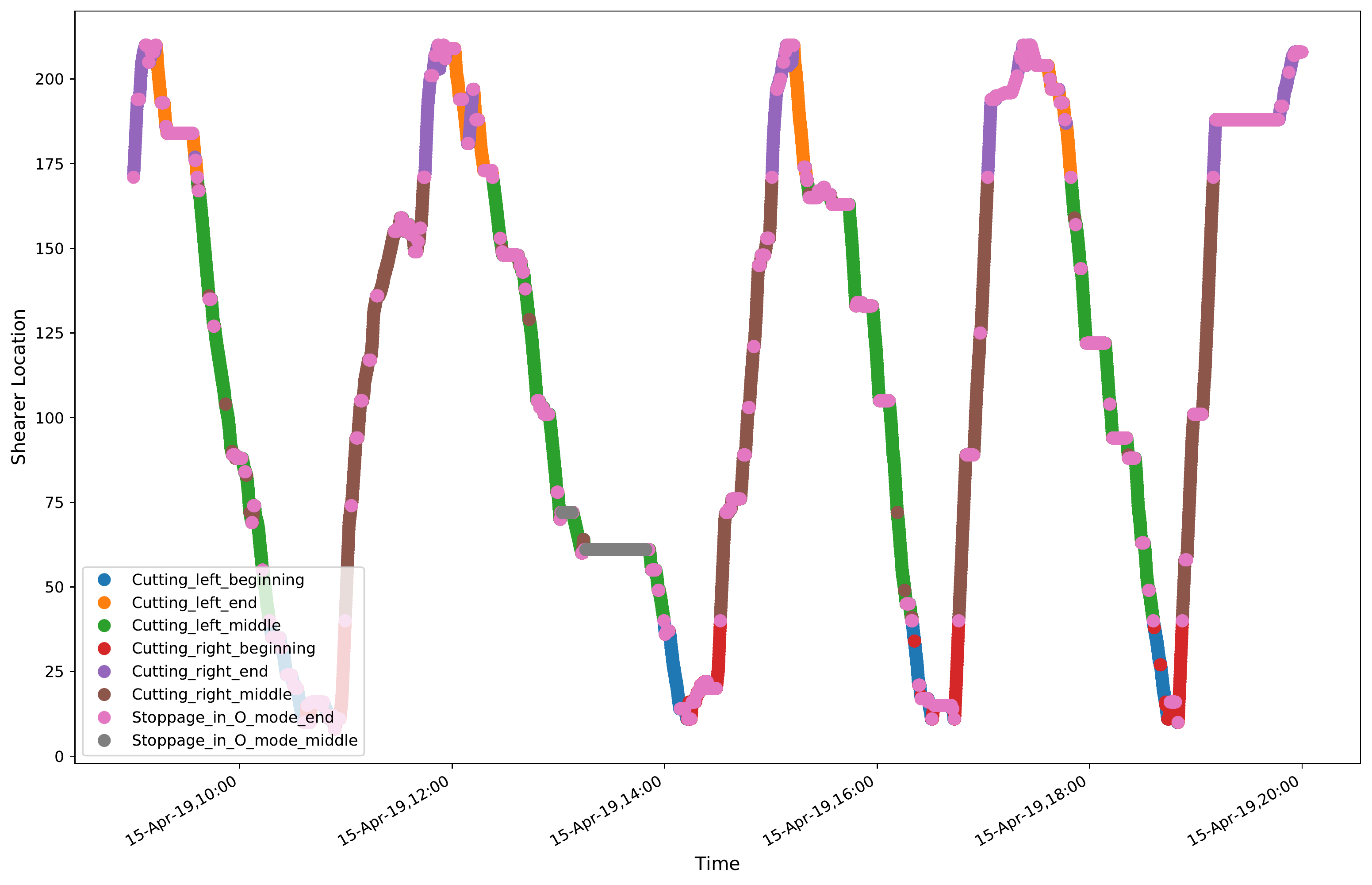}
\caption{Example of clusters discovered with expert knowledge.}
\label{fig:expert:stages}
\end{figure}

%

\subsection{Automated clustering}
Analysed sensor data is a multidimensional industrial log from the mining shearer. 
It contains 148 features that are raw sensor readings sampled every second. 
The full length of the data is about one year, and as a result, the analysed dataset contains almost 2 million time steps. 
In the analysis reported here, only a one-month time span indicated by domain experts as the most representative was taken into consideration.
As the analysed dataset contains real industrial data which is very often incomplete and noisy, there are many missing values.
Hence, the first step of the analysis was focused on data pre-processing. 

\subsubsection{Data pre-processing}

Most of the variables have more than 55\% of missing values and thus data pre-processing was started with data cleaning. 
The first step was to divide features into numeric, Boolean, and categorical. 
After this, columns that contained more than 65\% of missing data were removed. 
Finally, columns where the Pearson correlation coefficient was greater than 0.75 were removed unless a given feature was indicated by the domain expert as important. 

Among the features selected for further analysis, two subsets were created. 
The first subset contained features whose values can be interpolated and the second contained features for which values can be imputed. 
In the case of the former, a median strategy was applied to Boolean features and a mean strategy  to numeric features. 
For the features selected to be interpolated, the linear interpolation method was applied.

The last step of the dataset pre-processing was to create additional artificial features to obtain better performance from a clustering algorithm. 
In this step, the discretization of the selected features was applied. 
Features related to the electrical current (referring to shearer drums and haulages) were discretized in accordance with the following guidelines (based on the expert knowledge and nominal value given in the machinery documentation): 
\begin{itemize}
\item Idle value (0-10)\% of nominal value,
 \item Low load (10-40)\% of nominal value,
 \item Medium load (40-80)\% of nominal value,
 \item High load (80-100)\% of nominal value,
 \item Overload (above 100)\% of nominal value.
\end{itemize}

After applying all the steps described above, the dataset was correctly prepared as the input to the clustering algorithm.

\subsubsection{Clustering}

As the clustering method, we selected the deep temporal clustering algorithm~\cite{Madiraju2018}\footnote{Source code available under \url{https://github.com/FlorentF9/DeepTemporalClustering}}.
The selection of this method was dictated by the evaluation of several different clustering algorithms to select the one being the most promising and robust solution for our case.
This state-of-the-art algorithm integrates dimensionality reduction and temporal clustering into a single fully unsupervised end-to-end learning framework that allowed us to adapt this solution to the considered mining case. 

\begin{figure}[htb]
\centering
\includegraphics[width=\textwidth]{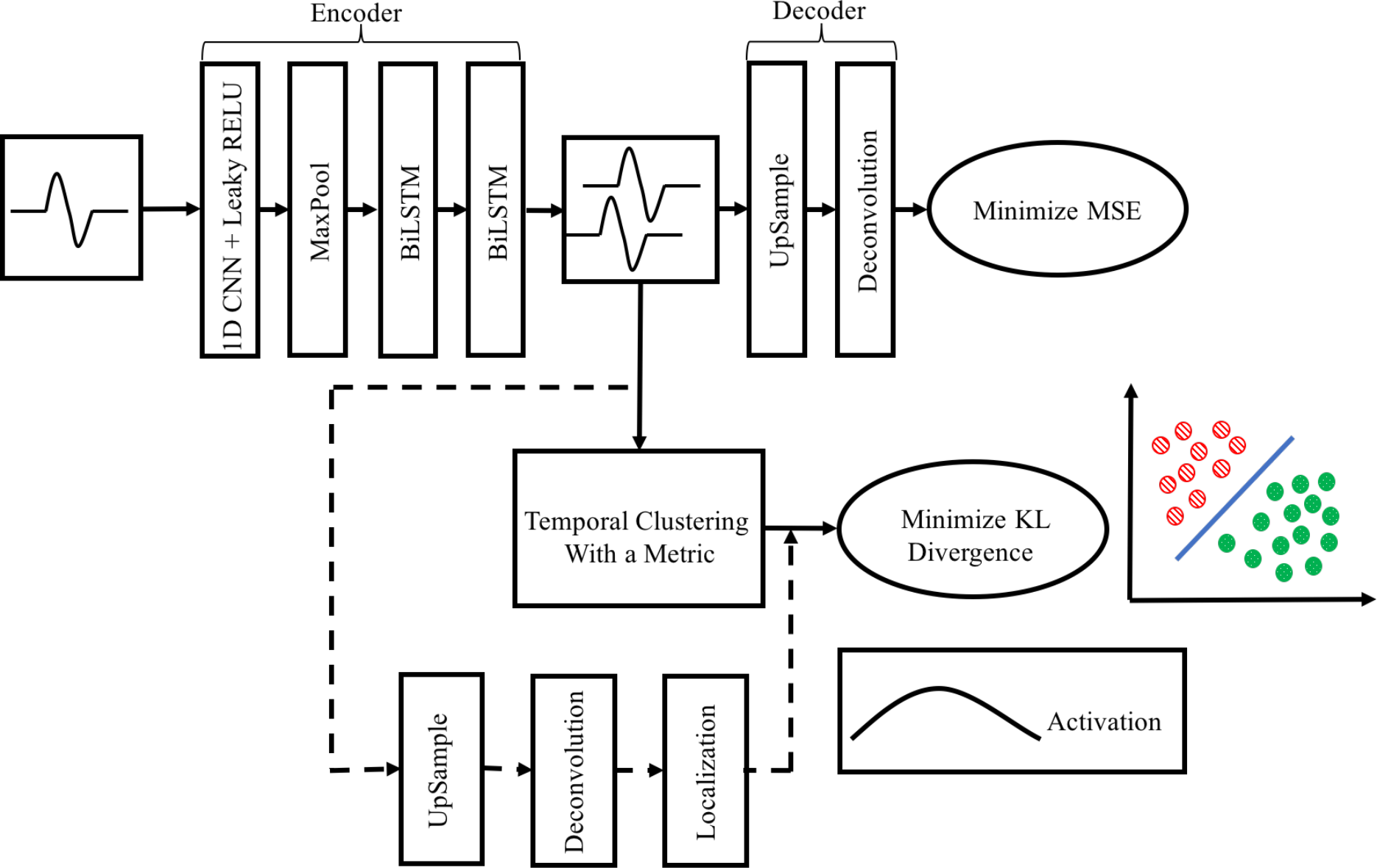}
\caption{Overview of a deep temporal clustering algorithm used~\cite{Madiraju2018}.}
\label{fig:DTC}
\end{figure}

The entire clustering algorithm is divided into two main parts. The first part concerns a temporal autoencoder (TAE) where an input signal is encoded into a latent space by a convolutional autoencoder followed by a BI-LSTM. The second part consists of a temporal clustering layer which generates the cluster assignments based on the latent representation of the BI-LSTM \cite{Madiraju2018}. The overview of the architecture of the Deep Temporal Clustering neural network is presented in Fig.~\ref{fig:DTC}.
The first level of the DTC algorithm consists of a 1D convolution layer followed by a max pooling layer and use of leaky rectifying linear units (Leaky ReLU). This approach allows for reducing the input data dimensionality and retaining only the most relevant information which is crucial for further steps as it helps to avoid long sequences. After this, the second level of the autoencoder is used which allows  obtaining  latent representation.
The use of BI-LSTM allows learning temporal changes in both time directions.
Finally, the last step is to assign a latent representation of the sequence $(x_0, x_1, \ldots, x_n)$, to the clusters. 
The algorithm uses two cost functions; the first cost function is provided by the mean square error (MSE) and is used to train the autoencoder. The use of this metric ensures that after applying dimensionality reduction, a good latent representation is sustained. Reconstruction is provided by an upsampling layer followed by a deconvolutional layer to obtain the autoencoder output. The second cost function is provided by the clustering metric -- the KL divergence. This approach ensures that the high-level features indeed separate the sequences $(x_0, x_1, \ldots, x_n)$ into $k$ clusters of distinct spatio-temporal behaviour. According to the obtained metrics, the algorithm modifies the weights in the BI-LSTM layer to optimally separate the sequences into clusters. 

The clustering layer consists of $k$ centroids $c_{i,j} \in (1, 2, \ldots, k)$. To initialise these cluster centroids, a latent signal $z_{i}$ was used. This latent signal was obtained by feeding the input $x_{i}$ through the initialised temporal autoencoder. A K-cut is performed to obtain the cluster and then average the elements in each cluster to get the initial centroids' estimates $c_{i,j} = (1, 2, \ldots, k)$.

After the estimation of the centroids $c_{j}$, the temporal clustering layer is trained using an unsupervised algorithm in two steps:
\begin{enumerate}
\item Compute the probability of assignment of input $x_{i}$ belonging to the cluster $j$. 
  The closer the latent representation $z_{i}$ of input $x_{i}$ is to the centroid $c_{i}$, the higher the probability of $x_{i}$ belonging to cluster $j$.    
  \item Update the centroid by using a loss function, which maximises the high confidence assignments using a target distribution $p$ Eq.~(\ref{eq:mat:tar_dist_kl}).
\end{enumerate}

To obtain cluster assignment, an input $z_{i}$ is fed to the temporal clustering layer and the distances $d_{i,j}$ are computed from each of the centroids $c_{j}$ using  similarity metrics.
The next step is to normalise the distance $d_{i,j}$ and convert it into probability assignments using a Student's t distribution kernel~\cite{LaurensvanderMaaten}.
\begin{equation}
    q_{i,j} = \frac{(1+\frac{siml(z_{i}, c_{j})}{\alpha})^{-\frac{\alpha+1}{2}}}{\sum^{k}_{J = 1}(1+\frac{siml(z_{i}, c_{j})}{\alpha})^{-\frac{\alpha+1}{2}}}
    \label{eq:mat:tar_dist}
\end{equation}

where:
\begin{itemize}
  \item $q_{i,j}$ is the probability of input $i$ belonging to cluster $j$, 
  \item $z_{i}$ corresponds to the signal in the latent space, obtained from the temporal autoencoder,
  \item $\alpha$ is the number of degrees of freedom of the Student's t distribution,
  \item $siml$ is the temporal similarity metric which is used to compute the distance between the encoded signal $z_{i}$ and centroid $c_{j}$.
\end{itemize}
In this study, similarity metrics were computed based on Euclidean distance.

To obtain the best results, the key is to train the temporal clustering layer iteratively. The main goal is to minimise the KL divergence loss between $q_{i,j}$ given by Eq.~(\ref{eq:mat:tar_dist}) and a target distribution $p_{i,j}$ given by Eq.~(\ref{eq:mat:tar_dist_kl}).
\begin{equation}
    p_{i,j} = \frac{q^{2}_{i,j}/f_{j}} {\sum_{j \in 1\ldots k}q^{2}_{i,j}/f_{j}}
    \label{eq:mat:tar_dist_kl}
\end{equation}
where:
$f_{j} = \sum^{n}_{i=1} q_{i,j}$. 
Further empirical properties were discussed in \cite{Xie2016}.
Based on this target distribution function, the KL divergence loss can be calculated using Eq.~(\ref{eq:mat:KL_loss}), where: $n$ and $k$ are the number of samples in the dataset and the number of clusters respectively.

\begin{equation}
    L = \sum_{i \in 1\ldots n}\sum_{j \in 1\ldots k}p_{i,j}\log \frac{p_{i,j}}{q_{i,j}}
    \label{eq:mat:KL_loss}
\end{equation}

To apply the described clustering algorithm, the first step was to prepare sliding window features based on the pre-processed dataset. 

Let us define a continuous multivariate time-series data $D$ of dimension $d$ with $n$ time-stamps, $D = (X_{1}, X_{2}, \ldots, X_{n})$ where each $X_{i} = \left \{x^{1}_{i},...,x^{d}_{i} \right \}$. 
Let $w$ be the window width, $s$ the stride, and $t$ the start time of a sliding window in the data.
A matrix $Z_k$ can now be defined, where each row is a vector of size $w$ of data extracted from the $k^{th}$ dimension~\cite{Ali2019}.
The matrix definition is given in Eq.~(\ref{eq:zmatrix}), where $r$ is the number of desired rows and the following inequality holds: $t+(r-1)s +w -1 \leq n$. 
In our work, the reprocessed dataset was divided into sliding windows containing $w=10$ time steps and stride set to $s=5$.

\begin{equation}
Z_{k}(w,s,t) = 
\begin{pmatrix}
x^{k}_{t} & x^{k}_{t+1} & \cdots & x^{k}_{t+w-1} \\
x^{k}_{t+s} & x^{k}_{t+s+1} & \cdots & x^{k}_{t+s+w-1} \\
\vdots  & \vdots  & \ddots & \vdots  \\
x^{k}_{t+(r-1)s} & x^{k}_{t+(r-1)s + 1} & \cdots & x^{k}_{t+(r-1)s + w-1} 
\end{pmatrix}
\label{eq:zmatrix}
\end{equation}

Having the input data prepared, the clustering algorithm was run with the following parameter values:
\begin{itemize}
\item The convolution layer has 10 filters with kernel size 5 and two Bi-LSTM’s have filters 50 and 1, respectively,
\item The pooling size is set to 5,
\item The deconvolutional layer has kernel size 5,
\item The autoencoder network is pre-trained using the Adam optimiser over 10 epochs,
\item Temporal clustering layer centroids are initialised using k-means clustering,
\item The mini-batch size is set to 512 for both pre-training and end-to-end fine-tuning of the network, 
\item Dropout is set to 0.1,
\item Recurrent dropout is set to 0.3.
\end{itemize}

The selected clustering algorithm assigns the same cluster to all time steps contained in a given window. 
Having the stride different than 0, there are multiple clusters assigned for a single time step. 
To determine the final cluster for a given timestep, the median of these assigned cluster numbers is calculated.  

The final clustering result, for a selected period of time, is presented in Fig.~\ref{fig:automated:stages}.
One can see that some parts of the plot are aligned with expert labelling presented in the previous section, but in some places, there are a lot more automatic clusters than those assigned by an expert.
Both expert labelling and automated labelling are inputs for \KNAC.

\begin{figure}[htb]
\centering
\includegraphics[width=\textwidth]{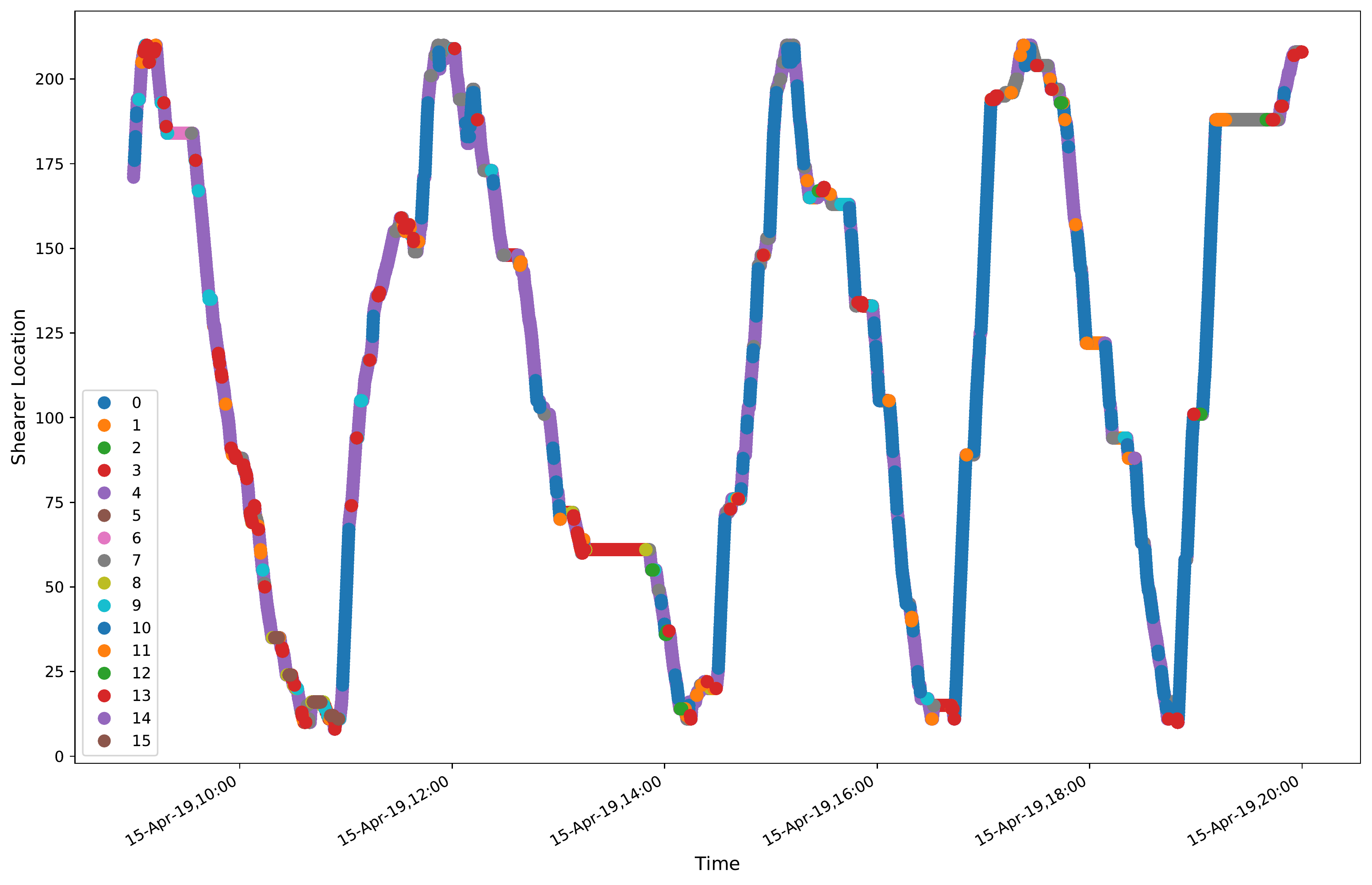}
\caption{Example of clusters discovered with DTC algorithm.}
\label{fig:automated:stages}
\end{figure}

\subsection{Knowledge refinement}
In this section, we describe the process of application of \KNAC to refinement of the initial expert knowledge.
The refinement was performed by an expert in the domain, who also delivered rules to encode prior knowledge and work in collaboration with Famur company, which delivered the data for analysis.
The input for this method is the contingency matrix depicted in Fig.~\ref{fig:ELvCL} which was computed with expert-based labelling and automated labelling, and partial results were given in Fig.~\ref{fig:expert:stages} and~\ref{fig:automated:stages}, respectively.

\begin{figure} [!ht]
\centering
\includegraphics[width=\linewidth]{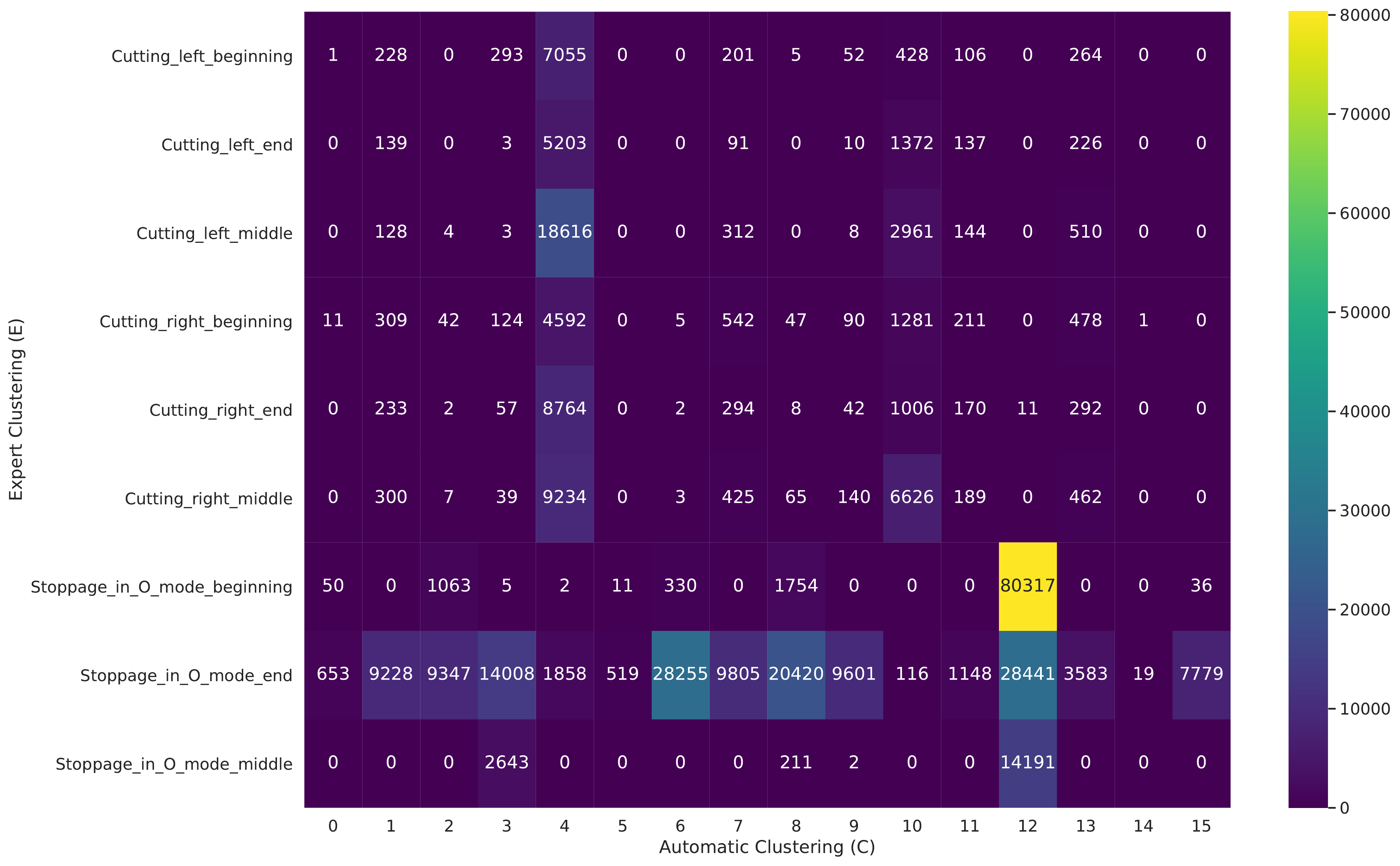}
\caption{Contingency matrix $M$ over expert clustering (rows) and automated clustering (columns).}
\label{fig:ELvCL}
\end{figure}

One can see that there are several  expert labels with various matched clusters, meaning that in the expert label, more specific states can be found (compared to what the expert expressed explicitly). 
Thus, these findings can be investigated in terms of tree rule extensions (Fig.~\ref{fig:tree}) with an assumption of a minimal confidence ratio threshold. 

We automated the process by providing two algorithms for \textit{split} and \textit{merge} operations, using the approach presented in Sec.~\ref{sec:knac}.
Below, the split suggestions are presented for a threshold value $\epsilon_s=0.8$ and silhouette weight $\lambda^{s}=0.5$ obtained from the $H^{split}$ matrix:

{\small
\begin{verbatim}
SPLIT Cutting_right_middle  INTO   [(4, 10)] (Confidence 0.64)
\end{verbatim}
}

In order to decide which clusters should be \textit{merged}, we used the $H^{merge}$ matrix and calculated cosine similarity between rows in the matrix.
As a result, we obtained a distance matrix that presents high cosine similarity between expert labels that were similarly split by the \textit{split matrix}.
Cosine similarity allowed us to bound the similarity between 0 and 1 and allow for a better comparison of cluster matches that differ in the number of points (magnitude of the vector).
The choice of expert labels to merge was parameterised by the threshold value.
Below, the merge suggestions are presented for a threshold value $\epsilon_s=0.8$ and linkage weight $\lambda^{m}=0.2$:

{\small
\begin{verbatim}
MERGE Cutting_left_middle	WITH Cutting_right_end	(Confidence 0.99)
MERGE Cutting_left_beginning WITH	Cutting_left_middle	(Confidence 0.99)
MERGE Cutting_left_end	 WITH   Cutting_left_middle	(Confidence 0.98)
MERGE Cutting_left_end WITH	Cutting_right_end	(Confidence 0.98)
\end{verbatim}
}

The aforementioned split and merge recommendation needs to be revised by an expert and incorporated into the knowledge base manually based on the explanations generated either as Anchor rules or decision stumps.

\subsection{Explanations}

We will focus only on one recommendation per type for the sake of simplicity.
In particular, we will investigate the split recommendation generated for the expert label \verb$Cutting_right_middle$ 
and the merge recommendation for two expert clusters, i.e.: \verb$Cutting_left_end$ and \verb$Cutting_right_end$.

\paragraph{Split explanations.}
The explanation for the recommendation on splitting the expert cluster \verb$Cutting_right_middle$  into clusters 4 and 10 returned by Anchor is given in Fig~\ref{fig:split_recoms}.
It was obtained using the XGBoost classifier.
In the split recommendation, some specific simpler states in the shearer operation can be found that are related to slower cutting (C\_4) and faster cutting  (C\_10). There is a clear explanation given by the \verb$SM_ShearerSpeed$ variable, indicating the slow movement of the shearer ($<$5.0) and fast movement ($>$10). 
Worth noting is that these simpler states can also be matched with cutting in other locations, as presented in the contingency matrix. 


\begin{figure}  
  \begin{subfigure}{1\linewidth}\label{fig:anchor_split}
    \centering
    \scriptsize{
    
    \begin{verbatim}C_4:                                            C_10: 
SM_ShearerSpeed_discrete <= 1.00                SM_ShearerSpeed > 10.00
AND SM_ShearerSpeed <= 5.00                     AND  RHD_EngineCurrent_discrete > 1.00 
AND LCD_AverageThree-phaseCurrent > 75.00    

Precision: 0.96 Coverage: 0.13                  Precision: 0.89 Coverage: 0.11\end{verbatim}
    }
\end{subfigure}

\begin{subfigure}{1\linewidth}\label{fig:dec-split-usecase}
\includegraphics[width=\textwidth]{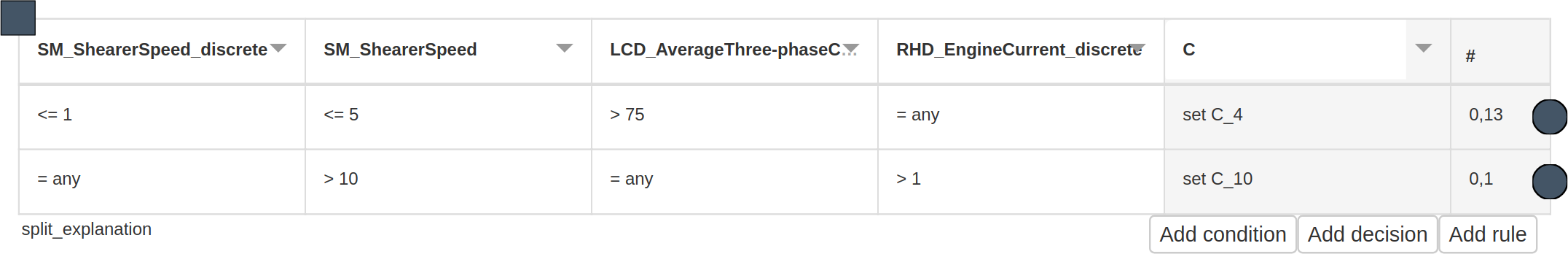}


\end{subfigure}
\caption{Explanations for the split recommendations with textual version obtained from Anchor at the top and visual representation in the form of \XTT table at the bottom.}
\label{fig:split_recoms}
\end{figure}

\paragraph{Merge explanations.}

Explanations for the recommendation on merging expert clusters \verb$Cutting_left_end$ and \verb$Cutting_right_end$ are given in Fig.~\ref{fig:merge_recoms}  for the Anchor explainer. 
Merging recommendations suggest joining two states of cutting realised at the end of the longwall face, with different movement directions. As a result, a more general state will be obtained. The suggested merge is relevant from a practical point of view. Very often, at a more general level of abstraction, such a state is isolated during shearer operation. What matters is this general state has its own specifics related to the location of operation. 



\begin{figure}
  \begin{subfigure}[b]{0.5\linewidth}\label{fig:anchor:merge}
    \centering
    \scriptsize{
\begin{verbatim}Cutting_left_end:               
SM_ShearerMoveInLeft > 0.00
Precision: 1.00 Coverage: 0.40

Cutting_right_end:  
SM_ShearerMoveInLeft <= 0.00
Precision: 1.00 Coverage: 0.60\end{verbatim}
}
\end{subfigure}
\begin{subfigure}[b]{0.5\linewidth}\label{fig:dec-merge-usecase}
\centering
\includegraphics[width=\textwidth]{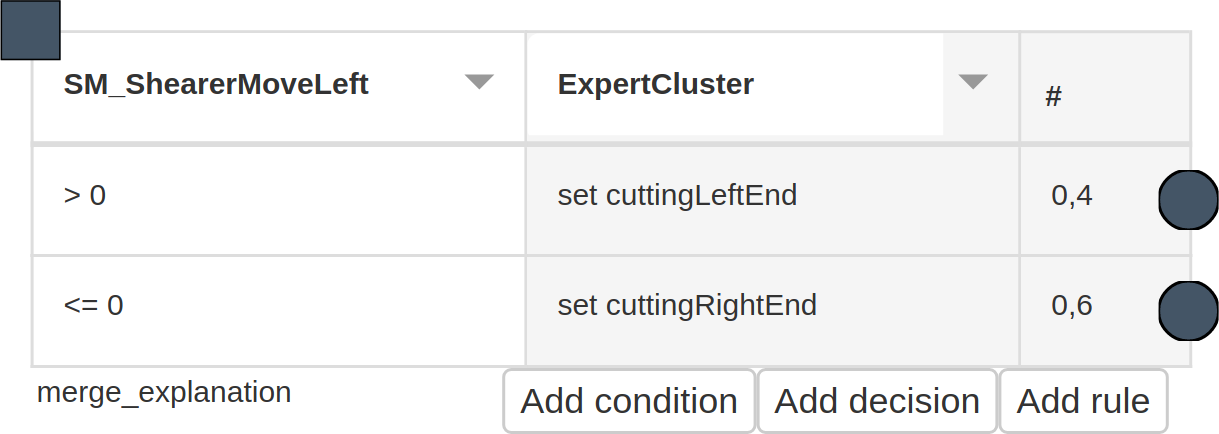}

\end{subfigure}
\caption{Explanations for the merge recommendations with textual version obtained from Anchor on the left and visual representation in the form of \XTT table on the right.}
\label{fig:merge_recoms}
\end{figure}

The final decision on the appropriate action is always left to the expert and, hence, additional analysis of the clusters that were pointed out by \KNAC can be performed with external methods.
However, this reduces the overhead of analysis only to the subset of clusters.


\section{Evaluation of the approach using an observational study}
\label{sec:eval}
In order to evaluate our approach on a larger group of experts, we designed a separate evaluation study that was based on the publicly available dataset from the e-commerce business area.
We used a sample dataset from Consumer Reviews of Amazon Products\footnote{See: \url{https://www.kaggle.com/datafiniti/consumer-reviews-of-amazon-products}, accessed (10.11.2021).}.
The sample consists of 1664 Amazon products with descriptions and one of the 26 categories assigned.
We treated categories from the dataset as expert labels and descriptions as data for clustering.
In Listing~\ref{lst:amazon}, a random product description from the dataset is presented for the category: \textit{Games \textgreater~Card Games}.
\begin{lstlisting}[label=lst:amazon,caption=Product description from the \textit{Games \textgreater~Card Games} category.,basicstyle=\scriptsize]
Eight types of creepy critters are depicted on the cockroach poker cards your 
job isn to win simply to make sure that you don lose  players take it in turns 
to pass card face down to one of their opponents and say out load what on the 
card they might bluff or they might be telling the truth  is it rat? or is it 
really stink bug? the recipient can either accept the card saying if they think
the player is telling the truth or bluffing  or they can peek at the card and 
pass it on box contains x rules in german english french italien  dutch
\end{lstlisting}

We intentionally corrupted the label assignment given in the dataset by performing fake merges and splits of several categories.
The goal of the study was to use \KNAC in order to find these corrupted categories of products. 
After the completion of the assignment, the experts were obliged to fill in the evaluation survey containing 13 questions:
\begin{enumerate}
\item How much time did you spend on the task?
\item Which Amazon categories should be merged into a meta-category? 
\item Which Amazon category should be split? Select all that should be split.
\item How confident are you about the correctness of your solution? 
\item How helpful were the recommendations provided by \KNAC for merges?
\item How helpful were the justifications of \KNAC recommendations for merges?
\item How helpful were the recommendations provided by \KNAC for splits?
\item How helpful were the justifications of \KNAC recommendations for splits?
\item How intuitive was usage of \KNAC?
\item How skilled are you in data science?
\item What is the biggest strength of \KNAC?
\item What is the biggest weakness of \KNAC?
\item Additional comments.
\end{enumerate}

We also gathered information about meta-parameters that participants finally selected for \KNAC merge and split recommendations.
This allowed us to reproduce their results and compare final clustering with classic, state-of-the-art approaches. 

The case study was available online through Google Colab Notebook and Microsoft Forms Survey\footnote{The study is available on the \KNAC GitHub repository: \url{https://github.com/sbobek/knac}.}. 
We tested three groups of target experts.
The first group contained five domain experts with knowledge about the e-commerce area.
The second group contained 3 data scientists with no professional knowledge of the e-commerce area.
The third group contained 22 participants with different, yet minimal experience in data-science and the e-commerce area.
Neither of the groups has prior experience with \KNAC.

\subsection{Setup of the study}
In the study, we assumed that the labelling provided by Amazon is correct and, therefore, we needed to corrupt the dataset to include some redundant categories (candidates for merges) and some general categories (candidates for split).
After the analysis of the dataset, we merged the  following two categories:
\begin{itemize}
\item \textit{Figures \& Playsets \textgreater~ Science Fiction \& Fantasy},
\item \textit{Characters \& Brands \textgreater~Star Wars \textgreater~Toys}.
\end{itemize}

Additionally, we split the category \textit{Puppets \& Puppet Theatres \textgreater~ Hand Puppets} into three synthetic categories:
\begin{itemize}
\item \textit{Figures \textgreater~ Fluffy},
\item \textit{Dolls \& Accessories \textgreater~ Teddybears},
\item \textit{Puppets \& Puppet Theatres \textgreater~ Hand Puppets}.
\end{itemize} 
We selected products for each of the synthetic categories randomly and distributed them uniformly between the three new categories.

On such data, we performed product description clustering in order to obtain automatic labels.
The initial pipeline containing data pre-processing, vectorization, dimensionality reduction, and clustering were provided for the users in order not to make the results dependent on the clustering method use or prior data prepossessing.
We followed a classic data clustering pipeline for the sake of simplicity and computational efficiency.
This included TF-IDF vectorization, SVD-based dimensionality reduction, and K-Means clustering.
We obtained 20 automatic clusters, compared to 26 clusters originally present in the dataset.

The experts were asked to use \KNAC to discover if there are clusters that should be merged or split.
The experts did not know that we had modified the original clustering and believed that they had been working on the real dataset.

Due to the fact that the dataset was not tabular data, but text, we changed the way the explanations were presented to the experts.
Figure~\ref{fig:knac:text:justification} presents the justification for a split for two arbitrary selected clusters which was generated with the LIME framework.
We also investigated translations from LIME format to rules if there was a need for an executable format of the new knowledge and the fulfilment of the requirements presented in Sec.~\ref{sec:orig}.
However, in the evaluation, we were mostly concerned with the recommendations and justification, hence, we did not include the aspect of the executable format.

\begin{figure}[htb]
\centering
\includegraphics[width=\textwidth]{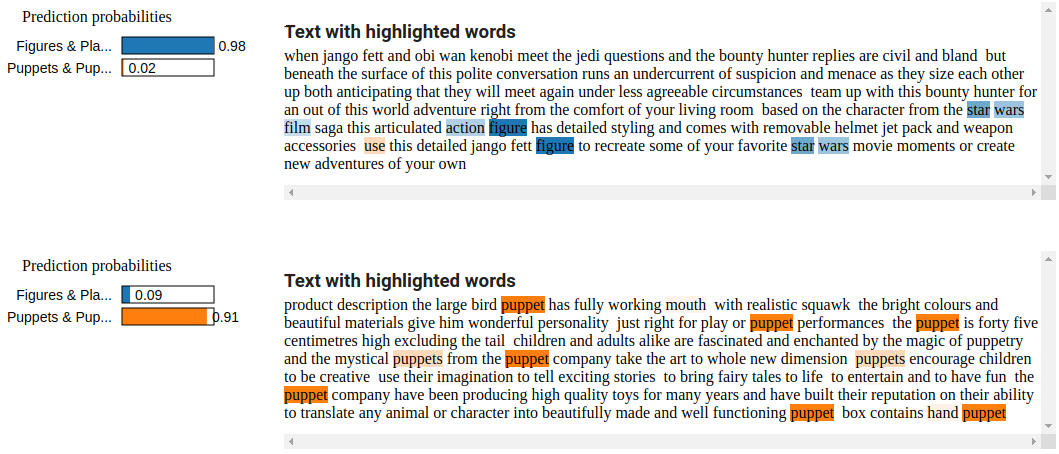}
\caption{Justification of a difference between two clusters obtained with LIME. Highlighted words were the most important in distinguishing the two clusters. The opacity of the highlight corresponds to the importance level.}
\label{fig:knac:text:justification}
\end{figure}

The \textit{blue} cluster represents \textit{Figures \& Playsets > Science Fiction \& Fantasy}, while the \textit{orange} cluster represents \textit{Puppets \& Puppet Theatres \textgreater~ Hand Puppets} categories.
The prediction probabilities indicate that there is high confidence in distinguishing one cluster from another (therefore, they should be separated concepts).
The importance of each word in favour for one of the clusters is depicted as the opacity of a highlight; the higher the opacity, the more important the word is.

\subsection{Discussion of results}


We have analysed the results in total for all of the participants, but also separately for each of the three groups: 1) e-commerce experts with data-science skills,  2) data-scientists without e-commerce skills, and 3) participants with neither experience in e-commerce nor data-science.
We first analysed how many of the artificially prepared categories were captured by the experts as merge and split candidates.
Over 93\% of the experts identified correctly at least two of the three categories which should be merged.
However, only 53\% of the participants correctly identified three categories as candidates for a merge. 
In the case of splits, over 53\% of participants correctly identified the category which we had combined previously.
Such a low number compared to the merge results may be caused by the fact that, in the categories we combined, split was very difficult to spot due to their similar content, and required a deeper look into the data, which either requires more data-science skills or domain knowledge.
The correct split identification rises to 71\% when we limit the observation to data-scientists and domain-experts only.

In addition, we compared final clustering results from automatic clustering algorithms and the \KNAC-guided approach with respect to completeness, homogeneity, and v-measure.
In Tab.~\ref{tab:knac_eval}, a comparison between \KNAC and other state-of-the-art approaches is shown.
For text vectorization, we used the clustering performed with TF-IDF vectorized text (Classic) and the clustering performed with transformer-based embedding (Transformer). 
For transformer-based embeddings, we used the all-MiniLM-L6-v2 pre-trained model trained on a corpus containing more than 1 billion sentences, available in the sentence-transformer\footnote{See: \url{https://www.sbert.net/}} package~\cite{reimers-2019-sentence-bert}.
The results presented in the Table were obtained from 12 participants who were given the same task as the previous group, however, \KNAC was never introduced to them and they were asked to perform clustering with classic approaches.
Participants had a choice to select the clustering algorithm they consider best for this task.
The selected algorithms included:  K-Means, Spectral clustering, Gaussian mixture, and Agglomerative clustering.

\begin{table}
\caption{Comparison of performance in clustering with \KNAC and other approaches. Mean values of the measure along with standard deviation are given.}
\label{tab:knac_eval}
\begin{tabularx}{\textwidth}{|X|X|X|X|}
\hline
 &  \textbf{Homogeneity} &  \textbf{Completeness} &  \textbf{V-measure} \\ \hline\hline
Classic & 0.70 $\pm$ 0.04 & 0.55 $\pm$ 0.04 & 0.62 $\pm$ 0.02\\ \hline
\KNAC & \textbf{0.98 $\pm$ 0.00} & \textbf{0.91 $\pm$ 0.02} & \textbf{0.94 $\pm$ 0.01}\\ \hline
Transformers & 0.75 $\pm$ 0.04 & 0.59 $\pm$ 0.05 & 0.66 $\pm$ 0.03\\ \hline
\end{tabularx}
\end{table}

As for their confidence in the results, e-commerce experts and regular participants  were more certain about the correctness of their answers than data-scientists, as depicted in Fig.~\ref{fig:conf}.
They give \KNAC loewr scores for intuitiveness than data-science experts, which may suggest they put more emphasis on their expert knowledge during the task resolution and were more sceptical towards \KNAC.
This observation was true for almost all of the survey questions, where \KNAC received lower scores from domain experts with lower data-science skills than from skilled data-scientists with no domain knowledge. 

\begin{figure}
  \begin{subfigure}{0.5\textwidth}
    \centering
\includegraphics[width=\textwidth]{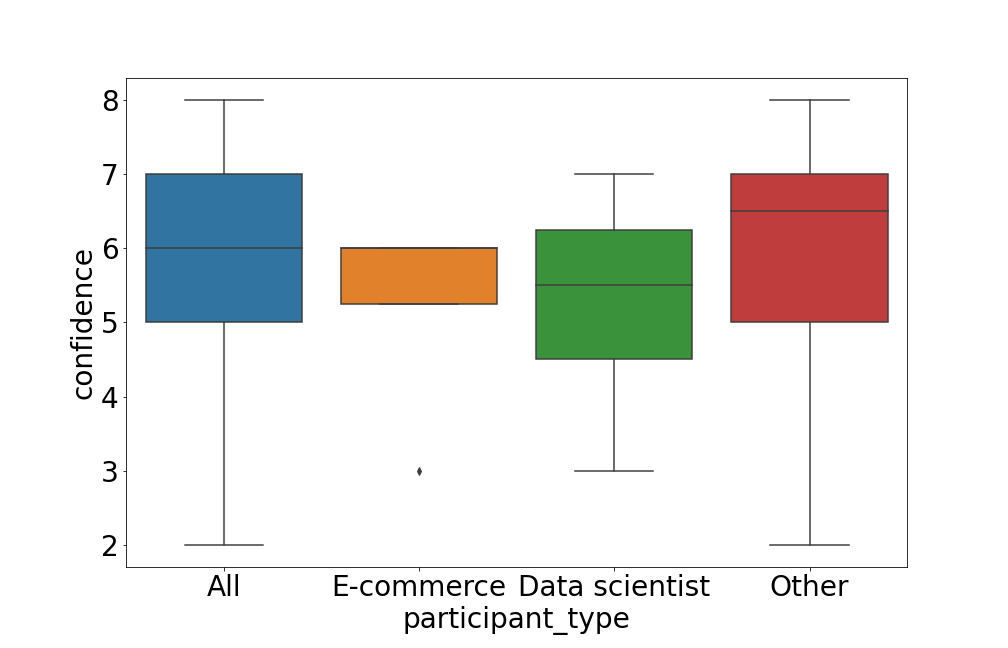}
\caption{Confidence in the solution. }
\label{fig:conf1}
  \end{subfigure}%
  \hspace*{\fill}   
  \begin{subfigure}{0.5\textwidth}
    \includegraphics[width=\textwidth]{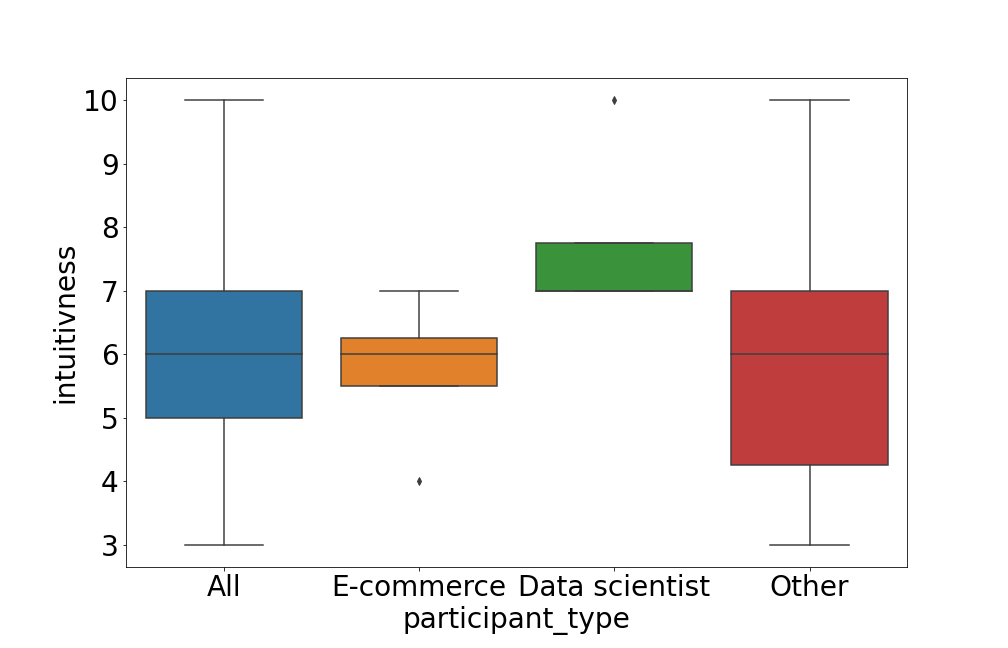}
\caption{Intuitiveness of \KNAC. }
\label{fig:conf}
  \end{subfigure}%
 
\caption{Confidence and intuitiveness of \KNAC by different groups of experts.} \label{fig:1}
\end{figure}

\begin{figure}
  \begin{subfigure}{0.5\textwidth}
    \centering
\includegraphics[width=\textwidth]{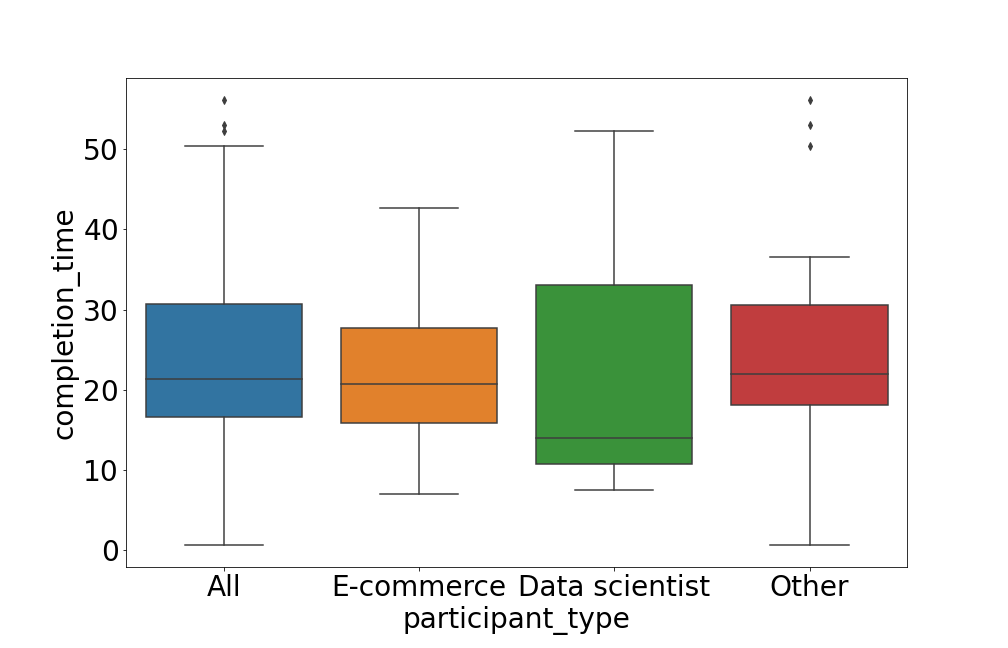}
\caption{Completion time. }
\label{fig:com:time}
  \end{subfigure}%
  \hspace*{\fill}   
  \begin{subfigure}{0.5\textwidth}
    \includegraphics[width=\textwidth]{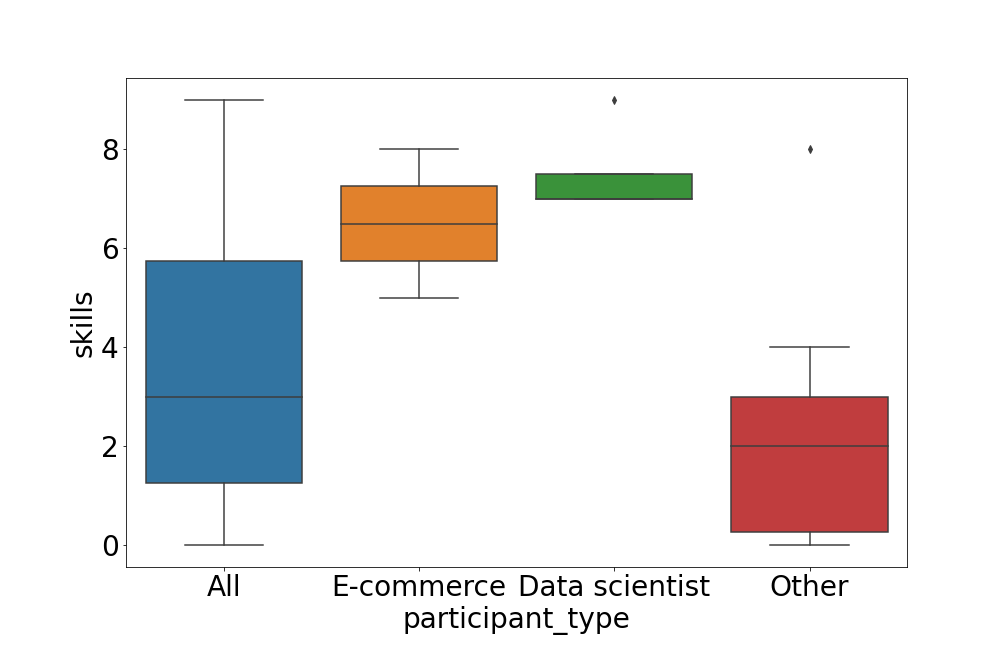}
\caption{Data-science skills. }
\label{fig:skills}
  \end{subfigure}%
 
\caption{Completion time of the task and their data-science skills.} \label{fig:1}
\end{figure}

\begin{figure}
  \begin{subfigure}{0.5\textwidth}
    \centering
\includegraphics[width=\textwidth]{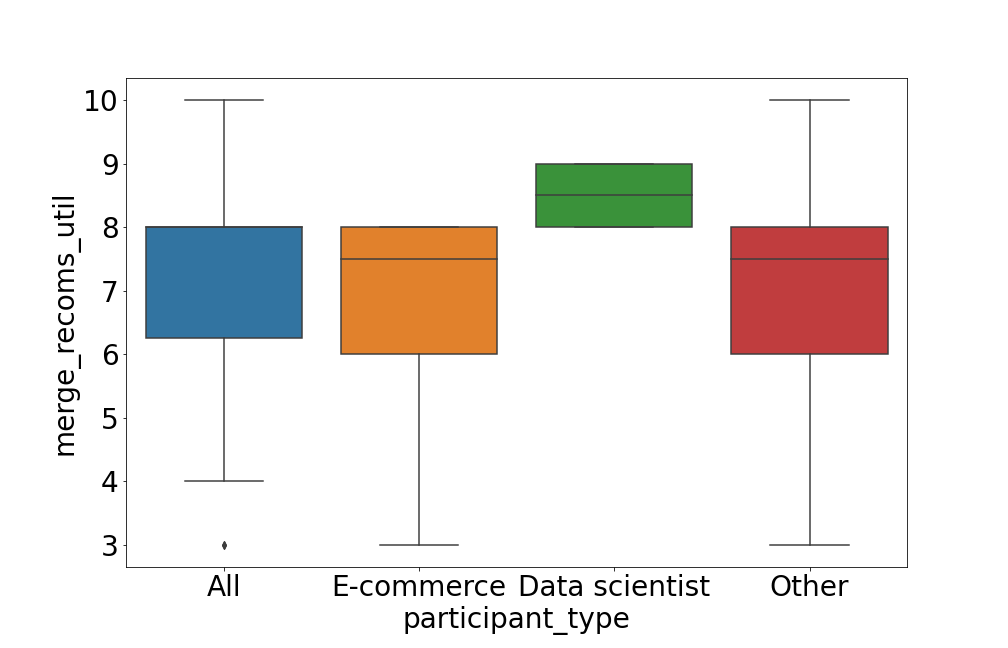}
\caption{Usefulness of the merge recommendations. }
\label{fig:merge1}
  \end{subfigure}%
  \hspace*{\fill}   
  \begin{subfigure}{0.5\textwidth}
    \includegraphics[width=\textwidth]{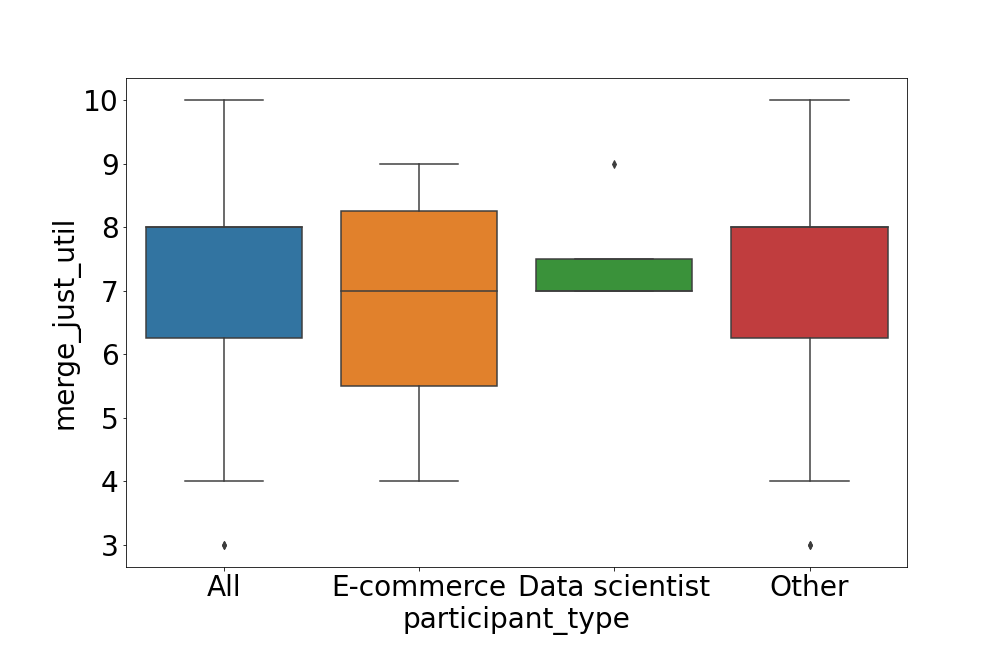}
\caption{Usefulness of merge justification. }
\label{fig:merge2}
  \end{subfigure}%
 
\caption{Usefulness of merge recommendations and justifications.} \label{fig:merge}
\end{figure}

Completion time for all of the groups was short and oscillated around 20 minutes, as presented in Fig.~\ref{fig:com:time}.
Groups with higher data-science skills spent less time on the task on average, as shown in Fig.~\ref{fig:skills}.
Merge recommendations and merge justifications were equally rated for all of the groups of experts (see Fig.~\ref{fig:merge}).
However, slightly better scores were assigned to \KNAC recommendations by data-scientists.
Split recommendations were given much worse ratings by e-commerce experts and participants without experience than by the data-science group of experts, as shown in Fig.~\ref{fig:split}.
This might be related to the specificity of the split we delivered, as the two categories that were combined by us as a candidate for a split were very similar in their content (Action figures and Star Wars characters toys), which might not be clear enough for the experts to divide them.
Also, the justifications for this case were not that obvious.

\begin{figure}
  \begin{subfigure}{0.5\textwidth}
    \centering
\includegraphics[width=\textwidth]{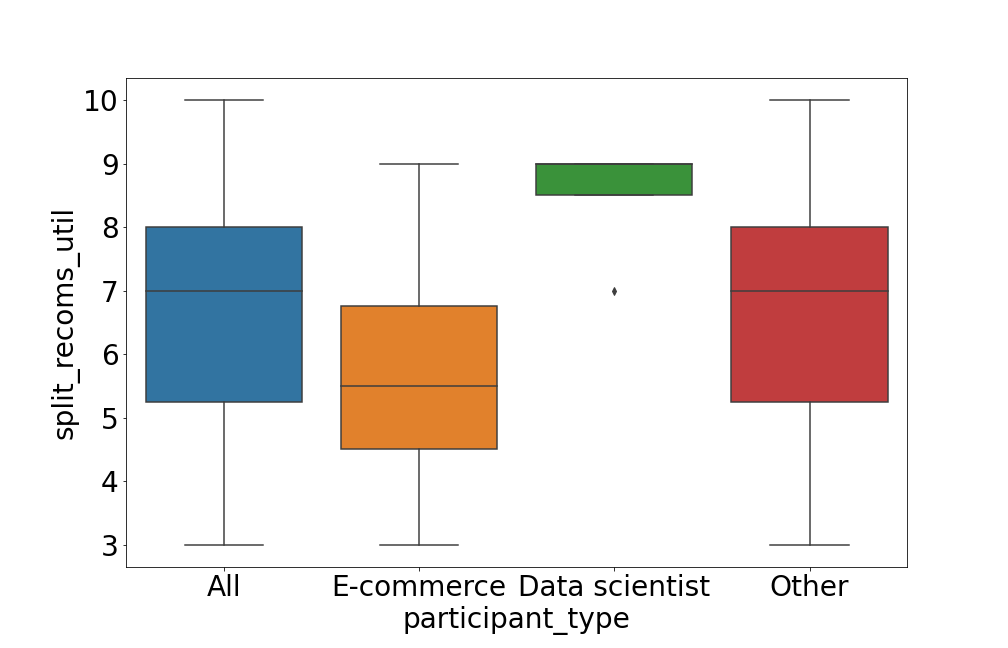}
\caption{Usefulness of split recommendations. }
\label{fig:split1}
  \end{subfigure}%
  \hspace*{\fill}   
  \begin{subfigure}{0.5\textwidth}
    \includegraphics[width=\textwidth]{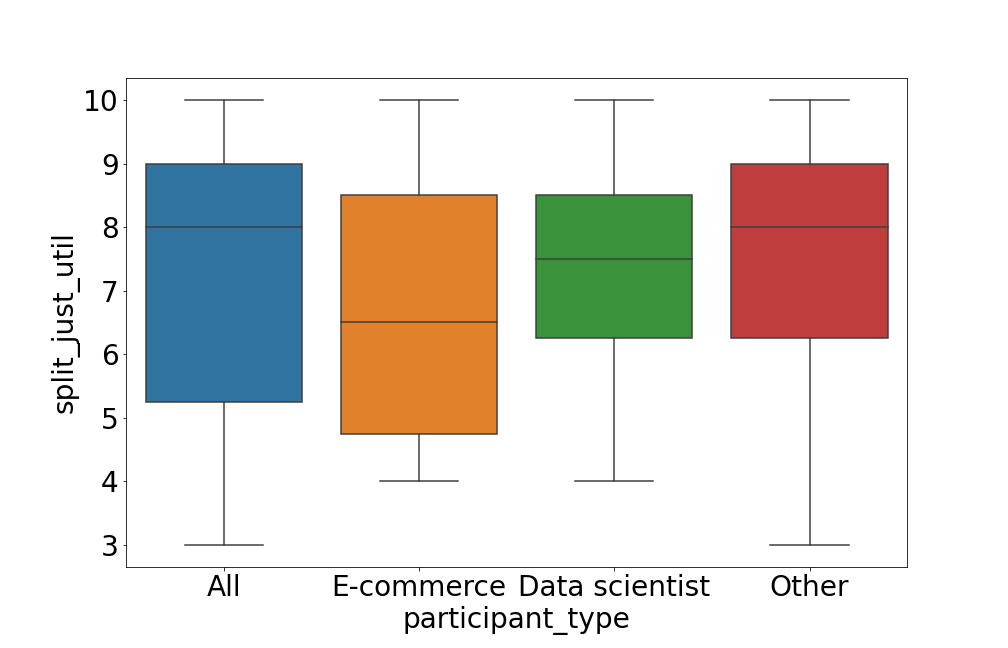}
\caption{Usefulness of split justification. }
\label{fig:split2}
  \end{subfigure}%
 
\caption{Usefulness of split recommendations and justification.} \label{fig:split}
\end{figure}

All of the participants  marked simplicity of the recommendation output as the biggest strength.
Some of the participants focused on the presentation aspects,  which was not the original contribution of \KNAC (we used the LIME method for text highlighting), yet this indicates how important the proper presentation of the explanation is for the audience. 
Finally, the biggest weakness was related to the process of fine-tuning (i.e., selection of threshold values), which was not intuitive for some of the experts and data scientists as well as the non-experienced participants.
This, however, does not undermine our results.

\section{Summary and future work}
\label{sec:summary}
In the paper, we introduced a new approach for extending the practical use of clustering algorithms for multidimensional time series.
\KNAC allows for the incorporation of expert knowledge into the clustering process, thus making it more robust and useful.
We presented the detailed formulation of our approach as well as demonstrated its practical use.
Using an industrial case study based on real-life sensor data, we described how an automated mechanism for labelling  operational states of an industrial device can be used to refine expert-based labelling.
These refinements were defined by us as \textit{splits} and \textit{merges} of expert labelling.
Such refined knowledge can contribute to a better understanding of the modelled phenomena.
As demonstrated in our case study, it can be used for generating more detailed reports on machine operational states as well as for detecting abnormal behaviour in the machinery, which was not detected by the original expert-knowledge rules.
To boost the reproducibility of our results and ease benchmarking, complementary to the real use-case scenario, we delivered artificially generated data sets with publicly available source code of \KNAC on GitHub~\footnote{Source code available under \url{http://github.com/sbobek/knac}}.
Moreover, to fully evaluate \KNAC as a generic method, we performed an evaluation study with the participation of experts on a public real-data set from a different domain (e-commerce).
The results we obtained from surveying the participants of the study indicate that both merge and split recommendations provided by \KNAC are useful in analysing clusters.
The quality of the clustering was much better with \KNAC than with classic and state-of-the-art clustering methods.
Furthermore, our approach delivers recommendations and suggestions along with explanations of these recommendations.
Both of these recommendations and explanations are encoded in the form of human-readable and executable rules.
Finally, our method is robust as it is able to work with arbitrary existing clustering algorithms.

Future works may include covering some of the limitations discussed below and wrapping the mechanism into a Python package available for installing from public repositories.
First of all, the order in which recommended splits and merges will be applied may affect the final clustering result and the final knowledge base.
This is especially important in cases when splits and merges concern the same expert clusters.
In the current version of our approach, there is no optimisation mechanism that could govern this process.

Furthermore, in our solution, we assume that the process is iterative, and hence it is safe to recommend minimal numbers of clusters for split/merge.
This is why for merges, only two cluster merges at a time are possible.
This may negatively affect the performance of the process and the granularity of the knowledge base.
For instance, in a case where there is a need to perform $n$ merges in order to converge, these merges will have to be performed independently $\lceil log2(n) \rceil$ times.

The final limitation that is currently under our investigation is the explainability mechanism used for explaining split and merge recommendations.
Most of the solutions use cluster centroids in the form of means or medians as representative examples of a cluster (cluster prototypes).
We also followed this trend in \KNAC.
However, this method has serious drawbacks in clusters that are of varying densities or shapes, so some other representation of clusters is required that will address this issue.
In fact, in our other research, we investigate the possible usage of multidimensional bounding boxes as  cluster representations for the explanation mechanism.
This will possibly contribute to the improvement of our method.

\section*{acknowledgements}
The paper is funded from the PACMEL project funded by the National Science Centre, Poland under the CHIST-ERA programme (NCN 2018/27/Z/ST6/03392).
The work of Szymon Bobek has been additionally supported by a HuLCKA grant from the Priority Research Area (Digiworld) under the Strategic Programme Excellence Initiative at the Jagiellonian University (U1U/P06/NO/02.16). 
The authors are grateful to ACK Cyfronet, Krakow for granting access to the computing infrastructure built in the projects No. POIG.02.03.00-00-028/08 "PLATON - Science Services Platform" and No. POIG.02.03.00-00-110/13 "Deploying high-availability, critical services in Metropolitan Area Networks (MAN-HA)"

\bibliographystyle{spmpsci}      

\end{document}